%% file: main.tex
\documentclass[sigconf, nonacm, pdfa]{acmart}
\usepackage{tikz}
\usepackage{url}
\usepackage{amsmath}
\usepackage{amsthm}
\usepackage{xcolor}
\usepackage{mathrsfs}
\usepackage{tablefootnote}
\usepackage{filecontents}
\usepackage{algpseudocode}
\usepackage{algorithmicx}
\usepackage{algorithm}
\usepackage{upgreek}
\usepackage{amsmath}
\usepackage{subcaption}
\usepackage{comment}
\usepackage{soul}
\usepackage{hyperref}
\usepackage{cleveref}
\usepackage{url}
\usepackage{booktabs}
\usepackage{multirow}
\usepackage{enumerate}
\usepackage{graphicx}
\usepackage{xspace}
\usepackage{framed}
\usepackage{enumitem}
\usepackage{marginnote}
\input{math_commands}

\input{macros}

\definecolor{revisioncolor}{RGB}{20,99,255} 

\newcommand{\rvspace}[1]{{\vspace{#1}}}

\newcommand{\revision}[1]{{#1}}
\newcommand{\margin}[1]{{}}
\newcommand{\marginoffset}[2]{{}}

\newcommand{\david}[1]{{\color{red} [David: ``#1'']}}

\usepackage{framed}
\usepackage{setspace}

\newcommand\vldbdoi{10.14778/3648160.3648184}
\newcommand\vldbpages{1473 - 1486}
\newcommand\vldbvolume{17}
\newcommand\vldbissue{6}
\newcommand\vldbyear{2024}
\newcommand\vldbauthors{Kezhao Huang, Haitian Jiang, Minjie Wang, Guangxuan Xiao, David Wipf, Xiang Song, Quan Gan, Zengfeng Huang, Jidong Zhai, and Zheng Zhang}
\newcommand\vldbtitle{\shorttitle} 
\newcommand\vldbavailabilityurl{https://github.com/xxcclong/history-cache}
\newcommand\vldbpagestyle{empty} 

\usepackage{balance}
\begin{document}
\date{}
\title{FreshGNN: Reducing Memory Access via Stable Historical Embeddings for Graph Neural Network Training}

\author{Kezhao Huang$^\ast$}
\affiliation{
  \institution{Tsinghua University}\city{}\country{}
}
\email{hkz20@mails.}
\email{tsinghua.edu.cn}

\author{Haitian Jiang$^\ast$}
\affiliation{
  \institution{New York University}\city{}\country{}
}
\email{haitian.jiang@nyu.edu}
 
\author{Minjie Wang$^\dag$}
\affiliation{
  \institution{Amazon}\city{}\country{}
}
\email{minjiw@amazon.com}
 
\author{Guangxuan Xiao}
\affiliation{
  \institution{MIT}\city{}\country{}
}
\email{xgx@mit.edu}
 
\author{David Wipf}
\affiliation{
  \institution{Amazon}\city{}\country{}
}
\email{daviwipf@amazon.com}
 
\author{Xiang Song}
\affiliation{
  \institution{Amazon}\city{}\country{}
}
\email{xiangsx@amazon.com}
 
\author{Quan Gan}
\affiliation{
  \institution{Amazon}\city{}\country{}
}
\email{quagan@amazon.com}
 
\author{Zengfeng Huang}
\affiliation{
  \institution{Fudan University}\city{}\country{}
}
\email{huangzf@fudan.edu.cn}
 
\author{Jidong Zhai}
\affiliation{
  \institution{Tsinghua University}\city{}\country{}
}
\email{zhaijidong@}
\email{tsinghua.edu.cn}
 
\author{Zheng Zhang}
\affiliation{
  \institution{Amazon}\city{}\country{}
}
\email{zhaz@amazon.com}



\begin{abstract}
A key performance bottleneck when training graph neural network (GNN) models on large, real-world graphs is loading node features onto a GPU. Due to limited GPU memory, expensive data movement is necessary to facilitate the storage of these features on alternative devices with slower access (e.g. CPU memory).  Moreover, the irregularity of graph structures contributes to poor data locality which further exacerbates the problem.  Consequently, existing frameworks capable of efficiently training large GNN models usually incur a significant accuracy degradation because of the currently-available shortcuts involved.
To address these limitations, we instead propose \system, a general-purpose GNN mini-batch training framework that leverages a historical cache for storing and reusing GNN node embeddings instead of re-computing them through fetching raw features at every iteration.  Critical to its success, the corresponding cache policy is designed, using a combination of gradient-based and staleness criteria, to selectively screen those embeddings which are relatively stable and can be cached, from those that need to be re-computed to reduce estimation errors and subsequent downstream accuracy loss. When paired with complementary system enhancements to support this selective historical cache, \system is able to accelerate the training speed on large graph datasets such as \texttt{ogbn-papers100M} and \texttt{MAG240M} by  3.4$\times$ up to 20.5$\times$  and reduce the memory access by 59\%, with less than 1\% influence on test accuracy. 
\end{abstract}

\maketitle

\pagestyle{\vldbpagestyle}
\begingroup\small\noindent\raggedright\textbf{PVLDB Reference Format:}\\
\vldbauthors. \vldbtitle. PVLDB, \vldbvolume(\vldbissue): \vldbpages, \vldbyear.\\
\href{https://doi.org/\vldbdoi}{doi:\vldbdoi}
\endgroup
\begingroup
\renewcommand\thefootnote{}\footnote{\noindent
This work is licensed under the Creative Commons BY-NC-ND 4.0 International License. Visit \url{https://creativecommons.org/licenses/by-nc-nd/4.0/} to view a copy of this license. For any use beyond those covered by this license, obtain permission by emailing \href{mailto:info@vldb.org}{info@vldb.org}. Copyright is held by the owner/author(s). Publication rights licensed to the VLDB Endowment. \\
\raggedright Proceedings of the VLDB Endowment, Vol. \vldbvolume, No. \vldbissue\ %
ISSN 2150-8097. \\
\href{https://doi.org/\vldbdoi}{doi:\vldbdoi} \\
}\addtocounter{footnote}{-1}\endgroup

\ifdefempty{\vldbavailabilityurl}{}{
\vspace{.3cm}
\begingroup\small\noindent\raggedright\textbf{PVLDB Artifact Availability:}\\
The source code, data, and/or other artifacts have been made available at \url{\vldbavailabilityurl}.
\endgroup
}

\blfootnote{$^\ast$ Contributed equally. Work done during internship at AWS Shanghai AI lab.}
\blfootnote{$^\dag$ Corresponding author.}

\input{intro}

\input{background}

\input{overview}
\input{history}

\input{system}
\input{eval}
\input{related}

\input{conclusion}

\begin{acks}
This work is partially supported by National Key R\&D Program of China under Grant 2021YFB0300300, National Natural Science Foundation of China (U20A20226), NSFC for Distinguished Young Scholar (62225206).
\end{acks}

\balance

\bibliographystyle{plain}
\bibliography{reference}


\end{document}

%% file: math_commands.tex
\usepackage{amsmath,amsfonts,bm}

\def\eqref#1{equation~\ref{#1}}

\def\1{\bm{1}}

\DeclareMathAlphabet{\mathsfit}{\encodingdefault}{\sfdefault}{m}{sl}
\SetMathAlphabet{\mathsfit}{bold}{\encodingdefault}{\sfdefault}{bx}{n}

\newcommand{\R}{\mathbb{R}}



%% file: macros.tex
\newtheorem{prop}{Proposition}[section]

\newcommand\blfootnote[1]{
  \begingroup
  \renewcommand\thefootnote{}\footnote{#1}
  \addtocounter{footnote}{-1}
  \endgroup
}

\newcommand{\m}[1]{\mathcal{#1}}
\newcommand{\cred}{\textcolor{red}}

\newcommand{\system}{\text{FreshGNN}\xspace}
\newcommand{\para}[1]{\noindent\textbf{#1}}
\newcommand{\mcsr}{\text{CSR2}\xspace}

%% file: intro.tex
\rvspace{-10pt}
\section{Introduction}
\label{sec:intro}

Graphs serve as a ubiquitous abstraction for representing relations between entities of interest. Linked web pages, paper citations, molecule interactions, purchase behaviors, etc., can all be modeled as graphs, and hence, real-world applications involving non-i.i.d. instances are frequently based on learning from graph data. To instantiate this learning process, graph neural networks (GNN) have emerged as a powerful family of trainable architectures with successful deployment spanning a wide range of graph applications, including community detection~\cite{community-detect}, recommender systems~\cite{gnn-recsys}, fraud detection~\cite{fraud-detect}, drug discovery~\cite{drug} and more.
The  predictive performance of GNNs is largely attributed to their ability to exploit both entity-level features as well as complementary structural information or network effects via  \emph{message passing} schemes~\cite{messagepassing}, whereby updating any particular node embedding requires collecting and aggregating the embeddings of its neighbors. Repeatedly applying this procedure by stacking multiple layers allows GNN models to produce node embeddings that capture local topology (with extent determined by model depth) and are useful for downstream tasks such as node classification or link prediction.


Not surprisingly, the scale of these tasks is rapidly expanding as larger and larger graph datasets are collected.  As such, when the problem size exceeds the memory capacity of hardware such as GPUs, a workaround is required, with some form of mini-batch training being the most common~\cite{graphsage,clustergcn,shadowgnn,graphsaint}.  Similar to the mini-batch training of canonical i.i.d. datasets involving images or text, one full training epoch is composed of many constituent iterations, each optimizing a loss function using gradient descent w.r.t.~a small batch of nodes/edges. In doing so, mini-batch training reduces memory requirements on massive graphs but with the added burden of frequent data movement from CPU to GPU.  The latter is a natural consequence of GNN message passing, which for an $L$-layer model requires loading the features of the $L$-hop neighbors of each node in a mini-batch.  The central challenge of efficient GNN mini-batch training then becomes the mitigation of this \textbf{data loading bottleneck}, which otherwise scales exponentially with $L$; even for moderately-sized graphs this quickly becomes infeasible.



Substantial effort has been made to address the data loading challenges posed by large graphs using system-level optimizations, algorithmic approximations, or some combination thereof.  For example, on the system side, GPU kernels are used to efficiently load features in parallel or store hot features in a GPU cache \cite{pytorch-direct,gnntier,gnnlab}; however, these approaches cannot avoid  memory access to the potentially large number of nodes that are visited less frequently.

On the other hand, there are generally speaking two lines of work on the algorithm front. 
The first is based on devising sampling methods to reduce the computational footprint and the required features within each mini-batch.  Notable strategies of this genre include neighbor sampling~\cite{graphsage}, layer-wise sampling~\cite{fastgcn,ladies}, and graph-wise sampling~\cite{clustergcn,shadowgnn}.  However, neighbor sampling does not solve the problem of exponential growth mentioned previously, and the others may converge slower or to a solution with lower accuracy~\cite{distdgl}.
Meanwhile, the second line of work~\cite{vrgcn,gas,graphfm} stores intermediate node representations computed for each GNN layer during training as \textit{historical embeddings} and reuses them later to reduce the need for recursively collecting messages from neighbors.
Though conceptually promising and foundational to our work, 
as we will later show in \Cref{subsec:background-mini-batch}, these solutions presently struggle to simultaneously achieve \textit{both} high training efficiency \textit{and} high model accuracy when scaling to large graphs, e.g., those with more than $10^8$ nodes and $10^9$ edges.

To this end, we propose a new mini-batch GNN training solution with system and algorithm co-design for efficiently handling large graphs while preserving predictive performance. 
As our starting point, we narrow the root cause of accuracy degradation when using historical embeddings to the non-negligible accumulation of estimation error between true and approximate representations computed using the history.  As prior related work has no practical mechanism for controlling this error, we equip mini-batch training with a \textit{historical embedding cache} whose purpose is to \textit{selectively} admit accurate historical embeddings while evicting those likely to be harmful to model performance. 
In support of this cache and its attendant admission/eviction policy, we design a prototype system called 
\underline{F}reshGNN: \underline{R}educing m\underline{E}mory access via \underline{S}table \underline{H}istorical embeddings,
which efficiently trains GNN models on large-scale graphs with high accuracy.  In realizing \system, our primary contributions are as follows:
\begin{itemize}[leftmargin=*]
    \item We propose a novel  mini-batch training algorithm for GNNs that achieves scalability without compromising model accuracy.  This is accomplished through the use of a historical embedding cache, with a corresponding cache policy that adaptively maintains node representations (via gradient and staleness criteria to be introduced later) that are likely to be stable across training iterations.  Moreover, by design our algorithm judiciously balances the caching on GPU of both embeddings and raw node features to reduce IO costs. In this way, we can economize GNN mini-batch training while largely avoiding the reuse of embeddings that are likely to lead to large approximation errors and subsequently, poor predictive accuracy.

    
    \item We create the prototype \system system to realize the above training algorithm with efficient implementation of subgraph pruning and data loading for both single-GPU and multi-GPU hardware settings.
    \item We provide a comprehensive empirical evaluation of \system across common baseline GNN architectures, large-scale graph datasets, and hardware configurations. Among other things, the results demonstrate that \system can closely maintain the accuracy of non-approximate neighbor sampling (within 1\%) while training 3.4$\times$ up to 20.5$\times$ faster than state-of-the-art baselines.
\end{itemize}

%% file: background.tex
\section{Background and Motivation}
\label{sec:background}

\subsection{Graph Neural Networks}
Let $\m{G} = (\m{V}, \m{E})$ be a graph with node set $\m V$ and edge set $\m E\subseteq \m V \times \m V$, where $n = |\m V|$. Furthermore, let $A\in\{0,1\}^{n\times n}$ be the graph adjacency matrix such that $A_{uv}=1$ if and only if there exists an edge between nodes $u$ and $v$. 
Finally, $X\in \R^{n\times d}$ denotes the matrix of $d$-dimensional node features (i.e., each row is formed by the feature vector for a single node) while $Y\in \R^{n\times c}$ refers to the corresponding node labels with $c$ classes. 

Given an input graph defined as above, the goal of a GNN model is to learn a representation $h_v$ for each node $v$, which can be used for downstream tasks such as node classification or link prediction. This is typically accomplished via a so-called \textit{message passing} scheme~\cite{messagepassing}.  For the $(l+1)$-th GNN layer, this involves computing the hidden/intermediate representation 
 \begin{equation}
     \label{equ:GNN-layer}
     \rvspace{-10pt}
 \begin{split}
    h_v^{(l+1)}&=f_{W}^{(l+1)}\left(h_v^{(l)}, \left\{h_u^{(l)}:u\in \m{N}(v)\right\}\right) \\[-5pt]
     &= \phi_{W}^{(l+1)}\left(h_v^{(l)}, \mathop{\text{AGG}}_{u\in \m{N}(v)}\left(\psi_{W}^{(l+1)}\left(h_u^{(l)}, h_v^{(l)}\right)\right)\right),
\end{split}
\rvspace{-10pt}
\end{equation}


\noindent where $h_v^{(l)}$ and $\m{N}(v)$ are the embedding and neighbors of node $v$ respectively, with $h_v^{(0)}$ equal to the $v$-th row of $X$. Additionally, $\psi$ computes messages between adjacent nodes while the  operator $\operatorname{AGG}$ is a permutation-invariant function (like sum, mean, or element-wise maximum) designed to aggregate these messages.  Lastly, $\phi$  represents an update function that computes each layer-wise embedding.  Note that both $\phi$ and $\psi$ are parameterized by a learnable set of weights $W$. 
Within this setting, the goal of training an $L$-layer GNN is to minimize an application-specific loss function $\m{L}({H}^{(L)}, Y)$ with respect to $W$.  This can be accomplished via gradient descent as $W\leftarrow W -\eta \nabla_{W}\m{L}$, where $\eta$ is the step size.


\subsection{Difficulty in Training Large-Scale GNNs}

Graphs used in GNN training can have a large number of nodes containing high-dimensional features (e.g., $d \in \{100,\ldots,1000\}$)~\cite{ogb,gcnrec}. As a representative example, within the widely-adopted Open Graph Benchmark (OGB)~\cite{ogb,ogb-lsc}, the largest graph \texttt{MAG240M} has 240M nodes with 768-dimensional 16-bit float vectors as features (i.e. 350GB total size); real-world industry graphs can be much larger still.
On the other hand, as nodes are dependent on each other, full graph training requires the features and intermediate embeddings of all nodes to be simultaneously available for computation, which goes beyond the memory capacity of a single GPU.

In light of this difficulty with full graph training, the most widely accepted workaround is to instead train with stochastic mini-batches \cite{graphsage, gcnrec,p3,gnnlab,mariusgnn}.
At each iteration, instead of training all the nodes, a subset/batch are first selected from the training set.  Then, the multiple-hop neighbors of these selected nodes (one hop for each network layer) are formed into the subgraph needed to compute a forward pass through the GNN and later to back-propagate gradients for training. Additionally, further reduction in the working memory requirement is possible by sampling these multi-hop nodes as opposed to using the entire neighborhood~\cite{graphsage}. The size of the resulting mini-batch subgraph with sampled neighborhood is much smaller relative to the full graph, and for some graphs can be trained using a single device achieving competitive accuracy~\cite{gnndesignspace} and generality~\cite{graphsage}. Hence mini-batch training with \textit{neighbor sampling} has become a \textbf{standard paradigm} for large-scale GNN training.

Yet even this standard paradigm is limited by a significant bottleneck: loading the features of the sampled multiple-hop neighbors in each mini-batch, which still involves data movement growing exponentially with the number of GNN layers. 
In many cases, data movement can occupy more than 85\% of the total training time.

\subsection{Existing Mini-Batch Training Overhauls}
\label{subsec:background-mini-batch}

Both system-level and algorithmic approaches have been pursued in an attempt to alleviate the limitations of mini-batch training.

\para{System Optimizations.} GNNLab~\cite{gnnlab} and GNNTier~\cite{gnntier} cache the raw features of frequently visited nodes to GPU using metrics such as node degree, weighted reverse PageRank, and profiling. However, for commonly-encountered graphs exhibiting a power-law distribution~\cite{powerlaw}, most of the nodes will experience a moderate visiting frequency and hence, the feature cache is unlikely to reduce memory access to them. More generally, because the overall effectiveness of this approach largely depends on graph structure, consistent improvement across different graph datasets is difficult to guarantee. PyTorch-Direct~\cite{pytorch-direct} proposes to store node features in CUDA Universal Virtual Addressing (UVA) memory so that feature loading can be accelerated by GPU kernels. Even so, data loading remains a bottle-neck for PyTorch-Direct, occupying 66\% of the total execution time, as the data transfer bandwidth is still limited by CPU-GPU bandwidth.



\para{Broader Sampling Methods.} While neighbor sampling reduces the size of each mini-batch subgraph, it does not completely resolve recursive, exponential neighbor expansion. Consequently, alternative sampling strategies have been proposed such as layer-wise~\cite{fastgcn,ladies} and graph-wise~\cite{clustergcn,shadowgnn} sampling. However, the resulting impact on model accuracy is graph-dependent and prior evaluations~\cite{distdgl,gnnlab} on large graphs like \texttt{ogbn-papers100M} indicate that a significant degradation (over 10\% accuracy drop) may occur.

\para{Reusing Historical Intermediate Embeddings.}
The other line of algorithmic work~\cite{vrgcn,gas,graphfm} for revamping mini-batch training approximates the embeddings of some node set $\m{S}$ using their historical embeddings from previous training iterations. This involves modifying the original message passing scheme from \ref{equ:GNN-layer} to become
\begin{equation}
\begin{split}
\label{equ:GNN-layer-hist}
h_v^{(l+1)}&=f_W^{(l+1)}\left(h_v^{(l)}, \left\{h_u^{(l)}\right\}_{u\in \m{N}(v)}\right) \\[-3pt]
    &=f_{W}^{(l+1)}\left(h_v^{(l)}, \left\{h_u^{(l)}\right\}_{u\in \m{N}(v) \setminus \m{S}} \cup \left\{h_u^{(l)}\right\}_{u\in \m{N}(v) \cap \m{S}} \right) \\\\[-15pt]
    &\approx f_{W}^{(l+1)}\bigg(h_v^{(l)}, \left\{h_u^{(l)}\right\}_{u\in \m{N}(v) \setminus \m{S}} \cup \underbrace{ \left\{\bar{h}_u^{(l)}\right\}_{u\in \m{N}(v) \cap \m{S}}}_{\text{\scriptsize Historical embeddings}} \bigg).
\end{split}
\end{equation}
The above computation is mostly the same as the original, except that now the node embeddings from $\m{S}$ are replaced with their historical embeddings $\bar{h}_u^{(l)}$. A typical choice for $\m{S}$ is any node not included within the {selected seed nodes}, and after each training step, the algorithm will refresh $\bar{h}_v^{(l+1)}$ with the newly generated embedding $h_v^{(l+1)}$ (authentic embedding). Using historical embeddings avoids recursive visits to neighbor node features and the aggregation of neighbor embeddings as well as the corresponding backward propagation.  This reduces not only the number of raw features to load but also the computation  associated with neighbor expansion. Moreover, the underlying training methodology is compatible with arbitrary message passing architectures.

While promising, there remain two major unresolved issues with existing methods that utilize historical embeddings. First, by unselectively recording the history, they all require an extra storage of size $O(Lnd)$ for an $L$-layer GNN, an amount which can be even larger than the total size of node features ($O(nd)$), a significant limitation. Secondly, historical embeddings as currently used may introduce impactful estimation errors during training.  To help illustrate this point, let $\tilde h_v^{(l+1)}$ denote the approximated node embedding computed using \Cref{equ:GNN-layer-hist}. The estimation error can be quantified by $\|\tilde h_v^{(l+1)}-h_v^{(l+1)}\|$, meaning the difference between the approximated embeddings and the authentic embeddings computed via an exact message passing scheme. 

\begin{figure}[ht]
\centerline{\includegraphics[width=0.4\textwidth]{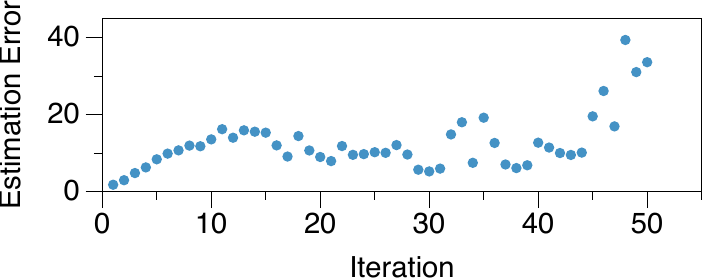}}
\caption{Average estimation error in one training epoch for GAS~\cite{gas} on \texttt{ogbn-products}.}
\label{fig:gas-error}
\end{figure}


\Cref{fig:gas-error} shows that the estimation error of each mini-batch increases considerably with more training iterations on the \texttt{ogbn}-\texttt{products} graph from OGB~\cite{ogb} when using GNNAutoScale (GAS)~\cite{gas}, a representative system based on historical embeddings. 
The root cause of this problem is that existing methods lack a mechanism for controlling the quality of the cached embeddings used for replacing message passing. Since the model parameters are updated by gradient descent after each iteration, the un-refreshed embeddings may gradually drift away from their authentic values resulting in a precipitous accuracy drop compared with the target accuracy from mini-batch training with vanilla/canonical neighbor sampling.

\begin{figure}[t]
\centerline{\includegraphics[width=0.45\textwidth]{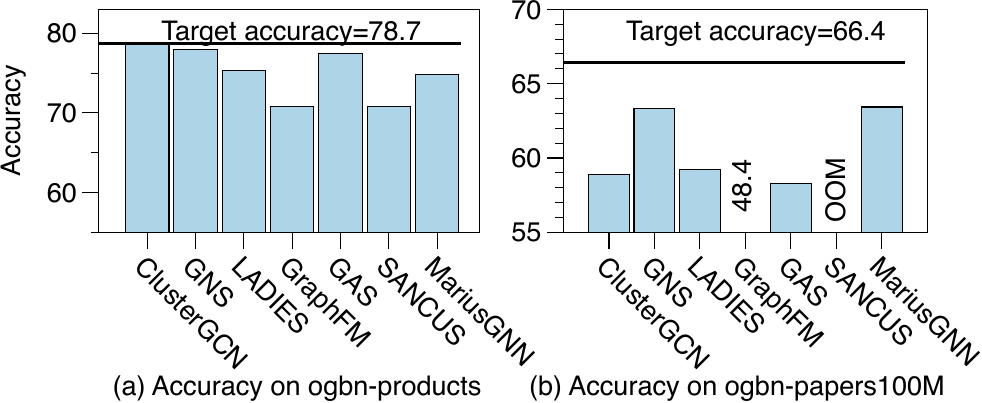}}
\caption{Test accuracy of mini-batch training algorithms (ClusterGCN~\cite{clustergcn}, GNS~\cite{gns}, LADIES~\cite{ladies}, GraphFM~\cite{graphfm}, GAS~\cite{gas}, and MariusGNN~\cite{mariusgnn}) and full graph training with historical embedding (SANCUS~\cite{sancus}) compared with the target accuracy achieved by (expensive) neighbor sampling on: (a) relatively small \texttt{ogbn-products} graph where the gap is modest for most algorithms, and (b) larger \texttt{ogbn-papers100M} graph where the gap grows significantly.}
\label{fig:teaser-acc}
\end{figure}


For the above reasons there remain ample room for new system designs that exploit historical embeddings for the setting of mini-batch training in a more nuanced way so as to maintain accuracy.  This is especially true as graph size grows larger and existing methods begin to exhibit accuracy degradation or run out of GPU memory as shown in \Cref{fig:teaser-acc} (see also Section~\ref{sec:eval-acc} for more in-depth evaluation of these approaches).

\subsection{Historical Embeddings in Full Graph Training}
\label{sec:history-full-graph}

{We note that the SANCUS algorithm from~\cite{sancus} also utilizes historical embeddings; however, the purpose is to reduce the communication overhead of multi-GPU \textit{full graph training} by refreshing the remotely cached embeddings when they drift away. Even so, SANCUS still relies on an $O(Lnd)$ storage for all the historical embeddings, making it run out of memory on large graphs like ogbn-papers100M, or else compromising the model capacity by the necessity of a small hidden size and/or lower floating precision. See \Cref{fig:teaser-acc} for representative examples and Section~\ref{sec:eval-acc} for further details. More critically, this approach requires access to the authentic embeddings themselves, which are naturally obtainable only in full graph training; in the setting of more scalable mini-batch training (our focus), computing them requires exact message passing, \textit{the very process we are trying to avoid}.}

%% file: overview.tex
\section{Design of \textbf{\system}}
\label{sec:overview}

In this work, we propose a new strategy for utilizing historical embeddings, with targeted control of the resulting estimation error to economize GNN mini-batch training on large graphs without compromising accuracy. Intuitively, this strategy is designed to favor historical embeddings with small $\|\bar h_u^{(l)}-h_u^{(l)}\|$ when processing each mini-batch, while avoiding the use of those that have drifted away. {However, directly computing this error requires the authentic embeddings $h_u^{(l)}$, which are only obtainable via the exact/expensive message passing we are trying to avoid as mentioned previously.} We therefore adopt an alternative strategy based on a key observation about the \textbf{stability} of the node embeddings during GNN mini-batch training: \textit{most of the node embeddings experience only modest change across the majority of iterations}.

\begin{figure}[ht]
\centerline{\includegraphics[width=0.35\textwidth]{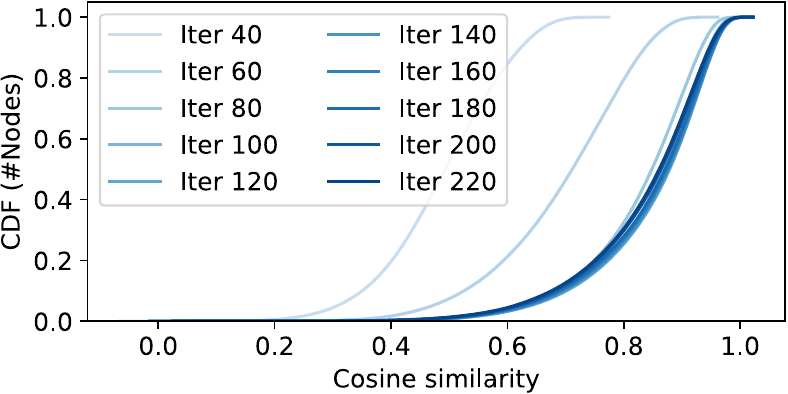}}
\caption{Distribution of cosine similarity between embeddings at iteration $t$ and embeddings at iteration $t-s$ during the training of a GCN model on \texttt{ogbn-arxiv}. Here $s=20$. }
\label{fig:cos-simlarity-emb}
\end{figure}

\para{Embedding Stability Illustration.} \Cref{fig:cos-simlarity-emb} showcases this phenomenon using a GCN~\cite{GCN} model trained on the \texttt{ogbn-arxiv} dataset. We measure the cosine similarity of the node embeddings at mini-batch iteration $t$ with the corresponding embeddings at $t-s$ for lag $s=20$ and plot the resulting distribution across varying $t$. After iteration 100 (the model converges with more than 500 iterations), most of the node embeddings exhibit a cosine similarity greater than 0.8.
This provides evidence of temporal stability in many node embeddings during GNN mini-batch training, which motivates a refined historical embedding cache to selectively detect, store, and reuse such stable node embeddings.

\input{algorithm/history_train}

\para{{\system} Training Algorithm.} To leverage the aforementioned embedding stability, \Cref{alg:history-train} introduces our mini-batch training process using the historical embedding cache. 
For each batch, we first generate a subgraph according to a user-defined sampling method (\Cref{line:train-sample}). A subgraph consisting of the sampled $L$-hop neighbors for the training nodes is returned, where $L$ is defined by whatever GNN model is chosen. Then for each layer, the nodes in the subgraph are divided into two types: normal nodes ($\m V_{normal}$ on \Cref{line:train-normal-nodes}) and those nodes whose embeddings can be found in the historical cache ($\m V_{cache}$ on \Cref{line:train-history-nodes}).  For the latter, embeddings are directly read from the cache (\Cref{line:train-fetch-history}), while for the normal nodes, embeddings are computed by the neighbor aggregation assumed by the GNN (\Cref{line:train-fetch-normal}). Finally, at the end of each iteration, the algorithm will utilize information from the forward and backward propagation steps to update the historical embedding cache. The goal is to check in stable embeddings that are more reliable for future reuse while checking out those that are not (\Cref{line:update}).

\begin{figure*}[t]
\centerline{\includegraphics[width=0.9\textwidth]{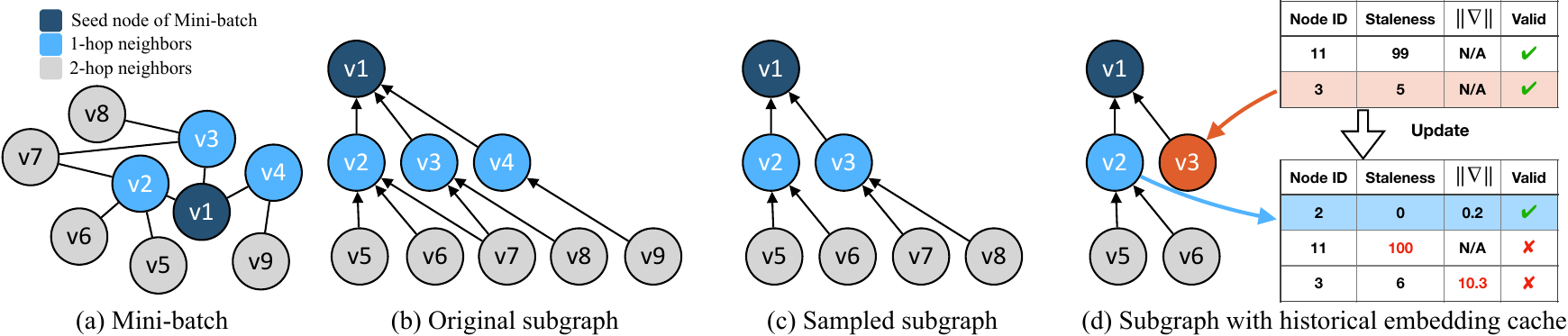}}
\rvspace{-10pt}
\caption{Illustration of historical embedding cache using an example mini-batch graph. }
\rvspace{-5pt}
\label{fig:overview}
\end{figure*}

\Cref{fig:overview} further elucidates \Cref{alg:history-train} using a toy example. Here node $v1$ is selected as the seed node of the current mini-batch as in \Cref{fig:overview}(a).  Computing $v1$'s embedding requires recursively collecting information from multi-hop neighbors as illustrated by the subgraph in \Cref{fig:overview}(b). As mentioned previously, neighbor sampling can reduce this subgraph size as shown in \Cref{fig:overview}(c). \Cref{fig:overview}(d) then depicts how the historical embedding cache can be applied to further prune the required computation and memory access. The cache contains node embeddings recorded from previous iterations as well as some auxiliary data related to staleness and gradient magnitudes as needed to estimate embedding stability. In this example, the embeddings of node $v3$ are found in the cache, hence its neighbor expansion is no longer needed and is pruned from the graph. Additionally, after this training iteration, some newly generated embeddings  (e.g., node $v2$) will be pushed to the cache for later reuse. Existing cached embeddings may also be evicted based on the updated metadata. In this example, both $v11$ (by staleness) and $v3$ (by gradient magnitude criteria to be detailed later) are evicted from the cache.


\begin{figure}[t]
\centerline{\includegraphics[width=0.48\textwidth]{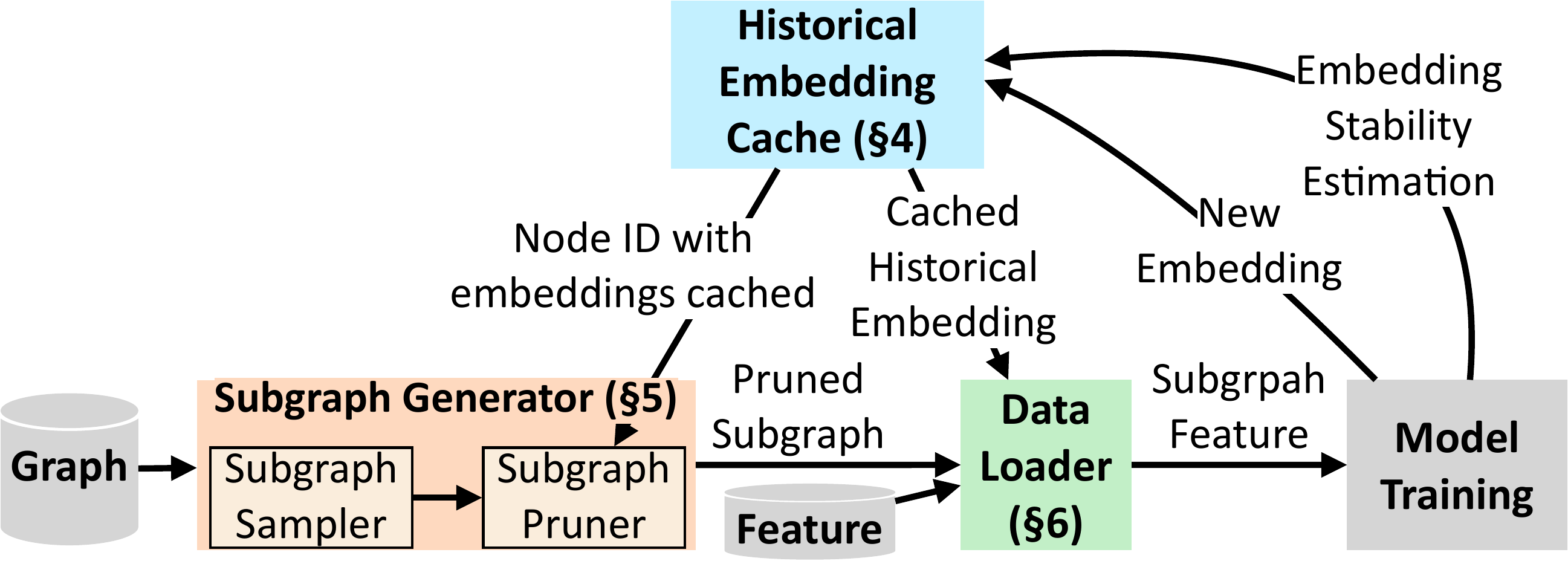}}
\rvspace{-5pt}
\caption{\system system workflow.}
\vspace{-10pt}
\label{fig:system-overview}
\end{figure}

\para{Main {\system} Components.}  To instantiate \Cref{alg:history-train}, and enable accurate, efficient GNN training over large graphs in CPU-GPU or multi-GPU scenarios, we require three system components, namely, (i) the historical embedding cache, (ii) the subgraph generator, and (iii) the data loader; each of these correspond with a colored block in \Cref{alg:history-train}).  \Cref{fig:system-overview} situates these components within a typical GNN training workflow, while supporting summaries are as follows (with subsequent sections filling in the full details):
\begin{enumerate}[leftmargin=*]
    \item The \textbf{historical embedding cache} is the central component of our \system design, selectively storing node embeddings in GPU and providing efficient operations for fetching or updating its contents as will be detailed in Section \ref{sec:cache}. 
    {Note that we also couple historical embeddings with frequently-visited raw features in the cache to reduce memory access.}
    \item The \textbf{subgraph generator} is responsible for producing a pruned subgraph given the current mini-batch and cached historical embeddings as will be discussed further in \Cref{sec:subgraph-generator}.  Compared with other GNN systems, a unique characteristic of our approach is that each subgraph structure is dependent on the nodes stored in the historical embedding cache, and the latter is dynamically updated at each training iteration. This essentially creates a reversed data dependency between the stages of mini-batch preparation and mini-batch training, making it difficult to apply pipelining to overlap the two stages like in other GNN systems~\cite{gnnlab,marius,mariusgnn}. To address this challenge, we further partition the workload into two steps: graph sampling and graph pruning, where only the latter step depends on the historical embedding cache. We then adopt a mixed CPU-GPU design that samples graphs in CPU while pruning graphs in GPU, 
    \margin{4-1}
    \revision{where CPU sampling can be overlapped by training time}; the pruning is complemented by a GPU-friendly data structure for fast graph pruning.
    
    \item The \textbf{data loader} is in charge of loading the relevant node features or historical embeddings given a generated subgraph. For each node that is not pruned by historical embeddings, the data loader fetches its raw features. Since these features are typically stored in a slower but larger memory device, \system further optimizes the data transmission for three scenarios: (1) fetching features from CPU to a single GPU, (2) fetching features from CPU to multiple GPUs, and (3) fetching features from other GPUs.  See Section \ref{sec:data-loader} for further details.
\end{enumerate}

%% file: algorithm/history_train.tex
\definecolor{bg1}{rgb}{1.0,0.85,0.75}
\definecolor{txt1}{rgb}{1.0, 0.66, 0.4}
\definecolor{bg2}{rgb}{0.76,0.94,0.78}
\definecolor{txt2}{rgb}{0.43, 0.84, 0.45}
\definecolor{bg3}{rgb}{0.76,0.94,1.0}
\definecolor{txt3}{rgb}{0.5, 0.8, 1.0}
\newcommand{\algbg}[3]{\colorbox{#1}{\parbox{#2\linewidth}{#3}}}

\begin{algorithm}[ht]
\setstretch{0.9}

\caption{Mini-Batch Training w/ Historical~Emb.~Cache }
\label{alg:history-train}
\begin{algorithmic}[1]
\State {{\bf Input}: Graph $\m{G} = (\m{V}, \m{E})$, input node features $H^{(0)}$, number of batches $B$, number of layers $L$, historical embedding cache $\m{C}$, rate for check-out using gradient $p_{grad}$, the maximum staleness threshold $t_{stale}$}
\State {$i \leftarrow 0$ \algorithmiccomment{\scriptsize Iteration ID}}
\State {$\{\m{B}_1, \cdots, \m{B}_B \} \leftarrow Split(\m{G}, B)$ \algorithmiccomment{\scriptsize Mini-batches of training nodes} }
\For {$\m{B}_b \in \{ \m{B}_1, \cdots, \m{B}_B \}$}
    \Statex \hspace{1.5em}\textcolor{txt1}{$\triangleright$ Subgraph generator}
    \State \algbg{bg1}{0.92}{$\m{G}_b \leftarrow sample(\m{G},\cup_{v\in \m{B}_b}) $} \label{line:train-sample}
    \vspace{-6pt}
    \For {$l \in \{ L-1, \cdots, 1 \}$} 
        \State \algbg{bg1}{0.85}{$\m{V}_{cache}^{(l)}\leftarrow \m{V}_{\m{G}_b}^{(l)} \cap \m{C}^{(l)}$ 
        \algorithmiccomment{\scriptsize History index lookup}} 
        \label{line:train-history-nodes}
        \vspace{-7pt}
        \State \algbg{bg1}{0.85}{$\m{G}_b.remove\_subgraph(root=\m{V}_{cache}^{(l)})$ \\
        \algorithmiccomment{\scriptsize Remove nodes for the calculation of $\m{V}_{cache}^{(l)}$}}
        \vspace{-5pt}
	    \State \algbg{bg1}{0.85}{$\m{V}_{normal}^{(l)}\leftarrow \m{V}_{\m{G}_b}^{(l)} \setminus \m{V}_{cache}^{(l)}$}
        \label{line:train-normal-nodes}
    \vspace{-5pt}
    \EndFor
    \Statex \hspace{1.5em}\textcolor{txt2}{$\triangleright$ Data loader}
    \State \algbg{bg2}{0.92}{$H_{normal}^{(0)}\leftarrow Load\_feature(\m V_{normal}^{(0)})$}
	\For {$l \in \{ 1, ..., L \}$} 
        \Statex \hspace{3em}\textcolor{txt3}{$\triangleright$ Historical embedding~cache}
    	\State \algbg{bg3}{0.85}{$h_{v}^{(l)} \leftarrow \m{C}^{(l)}[v], \forall v \in \m{V}_{cache}^{(l)}$}
        \label{line:train-fetch-history}
        \State {\thickmuskip=0mu $h_{v}^{{(l)}} \leftarrow  f_{W}^{(l)} \!\!\left(\!h_v^{(l-1)}\!,\left\{\!h_u^{(l-1)}\!\right\}_{\!u \in \m{N}_{\m{G}_b}\!(v)}\!\right)\!, \forall v \in \m{V}_{normal}^{(l)}$}
        \label{line:train-fetch-normal}
    \EndFor
    \State $loss = \m L(h^{(L)}_{\m{B}_b}, label_{\m{B}_b})$
    \State $loss.backward()$
    \State $i \leftarrow i \mathrel{+} 1$
    \For {$l \in \{ 1, \cdots, L \}$}
        \State \algbg{bg3}{0.85}{ \Call{Update}{$\m{C}^{(l)}, \m{V}_{normal}^{(l)} ,\m{V}_{\m{G}_b}^{(l)}, h^{(l)}, i, t_{stale}, p_{grad}$} }\label{line:update}
    \EndFor
    
\EndFor
\end{algorithmic}
\end{algorithm}

%% file: history.tex
\section{Historical Embedding Cache}
\label{sec:cache}


The historical embedding cache design is informed by the following: (1) \textit{What is a suitable cache policy for selecting \textbf{stable} node embeddings that favor high accuracy?} and (2) \textit{On top of this, how can we simultaneously take advantage of both embeddings and raw features to optimize system performance?}  We address each in turn below.

\subsection{Cache Policy for Accuracy}
Caching intermediate node embeddings is fundamentally different with caching raw node features. Unlike the raw node features staying unchanged, the embeddings are constantly updated during model training, meaning the quality of cached embedding will influence the accuracy of trained model. As a result, we have to selectively cache and reuse the stable embeddings.

\subsubsection{Caching Stable Embeddings}
In order to selectively cache historical embeddings, it is crucial to identify the stable ones. However, quantifying the stability of embeddings poses a significant challenge. In the context of GNN training, stability is measured by the disparity between a true node embedding and its corresponding cached version. A naive approach would involve recomputing all embeddings in the cache after each training iteration and removing those that have deviated significantly. However, this solution is impractical due to its reliance on computationally expensive data loading and computation, which contradicts the purpose of caching aimed at reducing costs.

\system introduces a lightweight approach utilizing \textit{gradient-based criteria} to identify stable embeddings. During training, the gradients of node embeddings are naturally computed to update the weight parameters, resulting in zero-cost acquisition of embedding gradients. These gradients serve as feedback from the model training process and can effectively indicate the stability of the embeddings. Specifically, a near-zero gradient magnitude suggests that the embedding contributes to accurate predictions and requires minimal adjustments during the current training iteration. In contrast, embeddings with large gradient magnitudes are considered less stable. By comparing the absolute values of embedding gradients, \system is capable of assessing the stability of embeddings at each iteration, enabling the storage of newly produced stable embeddings and invalidating unstable ones in the cache.

Based on the concept of using gradient as the indicator of embedding stability, we formulate the embedding cache policy for accuracy as follows.
Given a mini-batch graph, denote the set of nodes at layer $l$ as $\m{V}^{(l)}$ and the set of cached nodes as $\m{V}_{cache}$. 
For nodes $v\in \m{V}^{(l)}$, we use the magnitude of embedding gradients w.r.t.~the training loss as a proxy for node stability at each layer.  
\system admits nodes with small absolute values of gradients to the cache, with the rate controlled by $p_{grad}$, the fraction of newly generated embeddings to be admitted. Of the remaining $(1 - p_{grad})$ fraction of the nodes in the mini-batch, if any of these are already present in the cache, they will now be evicted.
%

In the example shown in \Cref{fig:cache-op}, at layer 1, Node 3 fetches its embedding from the cache while Node 2 computes its embedding faithfully by aggregating from  neighbors. During backward propagation, \system calculates both gradients $\nabla_{h_3^{(1)}} \m{L}$ and $\nabla_{h_2^{(1)}} \m{L}$, and compares their norms to decide which to admit or evict; here the embedding cache decides to admit Node 2 and evict Node 3.

\subsubsection{Evicting Stale Embeddings}

\begin{figure}[t]
\centerline{\includegraphics[width=0.35\textwidth]{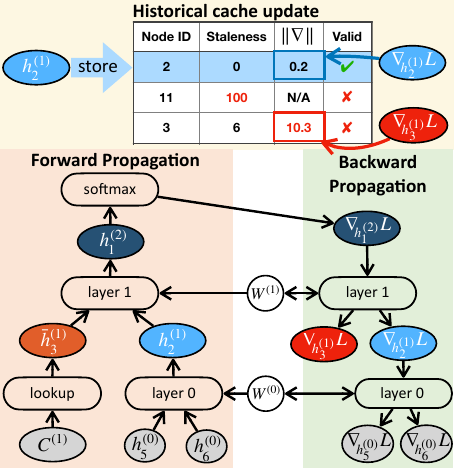}}
\rvspace{-5pt}
\caption{Illustration of the admission/eviction policy of the historical embedding cache. After a training iteration, the cache admits the embedding of Node 2 but evicts Node 3 (which was originally in the cache at the start of the iteration) by the gradient-based criteria. It also evicts Node 11 according to the staleness criteria.}
\rvspace{-10pt}
\label{fig:cache-op}
\end{figure}

To ensure model accuracy, it is essential to address the issue of stale embeddings that may arise due to the continuous weight parameter updates during each iteration. In addition to the gradient-based criteria, \system bounds the \textit{staleness} of the embeddings. The staleness is initially set to zero when an embedding is admitted to the cache. With each subsequent iteration, the staleness increases by one. \system considers embeddings with staleness exceeding a predefined threshold $t_{stale}$ as outdated and consequently evicts them from the cache. As illustrated in \Cref{fig:cache-op}, Node 11 is evicted based on this criterion. More broadly, utilizing and limiting staleness is a commonly used technique in training neural networks; the difference between \system and prior work that incorporates staleness criteria during training is discussed in \Cref{sec:related}.


\subsubsection{{Resulting Adaptive Cache Size}}

A notable difference from typical cache usage, which emerges from the above gradient-based and staleness-based criteria, is that our historical embedding cache size is not static or preset, but rather implicitly controlled by the two thresholds $p_{grad}$ and $t_{stale}$. Larger $p_{grad}$ or $t_{stale}$ means more embeddings to be cached but also requires the model to tolerate larger approximation errors introduced by historical embeddings. Moreover, setting $p_{grad}=0$ or $t_{stale}=0$ degrades the algorithm to normal neighbor sampling without caching historical embeddings. In contrast, setting $p_{grad}=100\%$ and $t_{stale}=\infty$ results in a policy that is conceptually equivalent to that used by GAS \cite{gas} and VR-GCN \cite{vrgcn}. So in this sense \system is a more versatile paradigm w.r.t.~previous historical embedding based methods. Later in Section \ref{sec:eval} we demonstrate that finding suitable values of $p_{grad}$ and $t_{stale}$ is relatively easy in practice; however, we leave open to future research on exploiting this flexible dimension within the historical cache design space.

\margin{4-2}
\revision{\subsubsection{Avoiding Initial Instability} \label{sec:instability}
According to \Cref{fig:cos-simlarity-emb}, we remark that many embeddings may be unstable at the beginning of training. To address this, \system can incorporate a \textit{stabilization period} when training begins. This involves initially disabling the historical cache and instead using regular mini-batch training. After the stabilization period, the historical cache is activated to reuse the stable embeddings. Experimental results in \Cref{sec:eval} demonstrate that introducing a short stabilization period (less than half of an epoch) can at times lead to improved model quality.}

\subsection{Cache Policy for System Performance}
\label{sec:gpu-implement}

The cache policy of \system, aimed at ensuring model accuracy, incorporates gradient and staleness criteria to maintain stable and up-to-date embeddings. 
However, it is also crucial to consider the cache's impact on system performance.
Therefore, in addition to accuracy considerations, we refine the cache policy by integrating both embeddings and raw features to minimize memory access and improve system performance. 

The design of a runtime cache should be capable of effectively managing both embeddings and features on the GPU becomes imperative. However, the dynamic nature of embeddings presents two distinct challenges.
The first challenge involves the generation of features and embeddings, noting that raw features remain static whereas embeddings are dynamically generated during training. Consequently, it is crucial to efficiently cache the embeddings while leveraging the static property of features, thereby optimizing resource utilization.
The second challenge revolves around the eviction of features and embeddings. Raw features can be retained in the cache permanently, whereas embeddings necessitate dynamic eviction to maintain accuracy. Therefore, it is important to invalidate evicted embeddings with minimal overhead while simultaneously freeing up space to accommodate new embeddings.

\begin{figure}[ht]
\rvspace{-5pt}
\centerline{\includegraphics[width=0.35\textwidth]{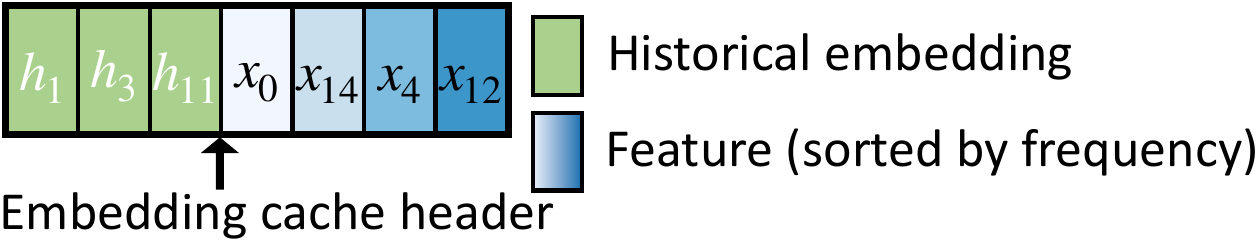}}
\rvspace{-5pt}
\caption{Bidirectional cache implementation.}
\rvspace{-5pt}
\label{fig:cache-impl}
\end{figure}

To tackle these challenges, we propose a novel solution called the \textit{bidirectional cache} to effectively manage both embeddings and features. The fundamental concept is to statically cache frequently visited features on one side of the buffer, while dynamically caching the embeddings generated during training on the other side. 
At the junction between embeddings and features in the buffer, features are less frequently visited and will be replaced by newly generated embeddings. 
The detailed design of the bidirectional cache is presented in three distinct parts: the feature side, the embedding side, and the junction where they converge.

Before the training process, \system populates the available GPU buffer with hot raw features (right hand side in \Cref{fig:cache-impl}). These features are ordered based on their estimated saved memory access, ensuring that those with larger saved memory access are positioned closer to the feature side of the buffer. This placement reduces the likelihood of these features being replaced by newly generated embeddings present on the other side of the buffer.

During the training phase, \system dynamically stores the newly generated embeddings on the embedding side of the buffer (left hand side in \Cref{fig:cache-impl}). To optimize runtime efficiency, GPU threads are utilized to effectively manage embeddings in parallel. To facilitate this parallel management, \system maintains a node ID mapping array with a length of $O(|\m{V}|)$, where each entry stores the index of the cache that holds the embedding of the corresponding node. The additional storage required for this mapping array is affordable in comparison to the storage of the embeddings themselves. Moreover, this mapping array ensures that the compute complexity of each fetching operation is $O(1)$. Since there are no dependencies between different entries, \system can fully exploit the parallelism capabilities of the GPU.

At the junction of cached features and embeddings, \system employs a ring buffer design that effectively admits new embeddings while evicting old ones. This design ensures that features are not unnecessarily displaced by newly generated embeddings. As illustrated in \Cref{fig:cache-impl}, \system maintains an embedding cache header. It initially points to the first item of the cache and moves forward when new embeddings are added. At every $t_{stale}$ iteration, \system resets the embedding cache header to the beginning. Consequently, newly added embeddings overwrite outdated ones, naturally evicting them from the cache. To evict embeddings with significant gradient magnitudes, \system employs an approach where it invalidates the corresponding entries in the node ID mapping array, instead of physically deleting them. These invalidated slots are naturally recycled within the ring buffer design.

%% file: system.tex


\section{Cache-Aware Subgraph Generator}
\label{sec:subgraph-generator}

In \system, the mini-batch subgraphs are generated adaptively according to the node embeddings stored in the cache. As the cache is in turn updated at the end of each iteration, this prevents \system from adopting a naive pipelining strategy to parallelize subgraph generation and model training as in other systems. To address this challenge, we decompose subgraph generation into two steps: graph sampling and graph pruning, where only the latter depends on the historical embedding cache. We then adopt a mixed CPU-GPU design to further accelerate them.

\para{Asynchronous CPU Graph Sampling.}
This step first extracts/samples subgraphs normally for the given mini-batch and then moves them to GPU without querying information from the cache. As a result, graph sampling can be conducted asynchronously with the later GPU computation. We further utilize multithreading instead of multiprocessing to produce multiple subgraphs concurrently in contrast to the existing systems like DGL~\cite{dgl} and PyG~\cite{pyg}. Additionally, we use a task queue to control the production of subgraphs and avoid overflowing the limited GPU memory.

\begin{figure}[h]
\centerline{\includegraphics[width=0.48\textwidth]{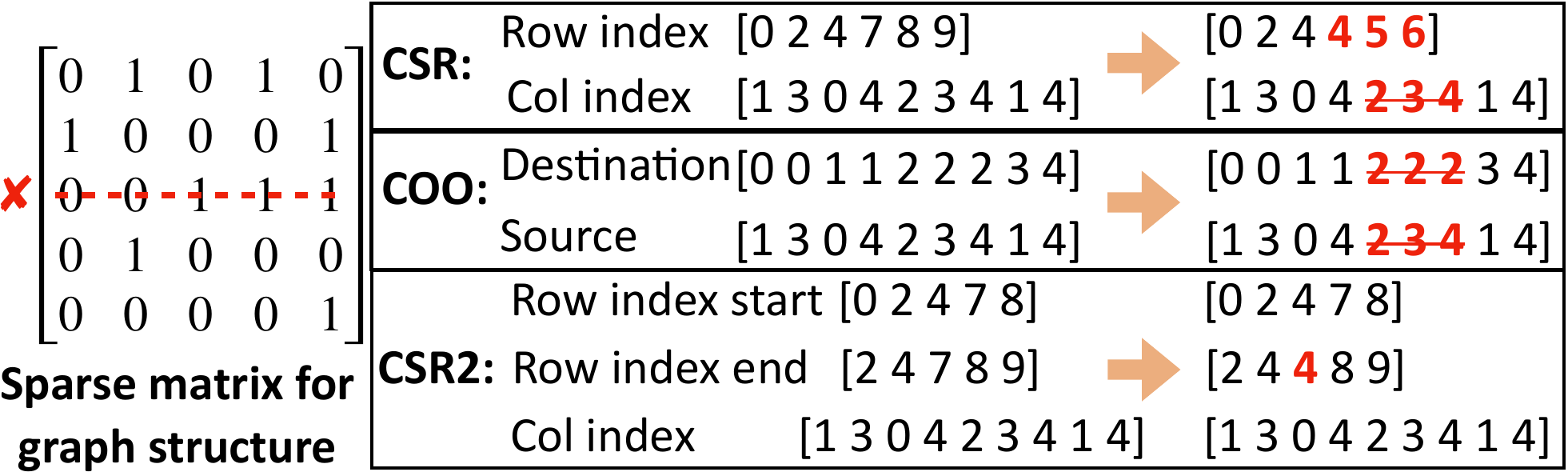}}
\caption{Removing a center node's neighbors using different graph data structures}
\rvspace{-5pt}
\label{fig:sparse-format}
\end{figure}

\para{GPU Graph Pruning.}
The graph pruning step scans the mini-batch graph from the seed node layer to the input node layer. For any cached node, it recursively removes all the multiple-hop neighbors so that the corresponding computation is no longer needed for model training. The remaining challenge is that traditional sparse formats are not suitable for parallel modification in GPU. As shown in \Cref{fig:sparse-format}, the Coordinate (COO) format represents a graph using two arrays containing the source and destination node IDs of each edge. 
{Pruning incoming neighbors of a node in COO requires first locating neighbors using a binary search and then deleting them from both arrays. The prune complexity is $O(\log(|\m{E}_{sample}|) + N_{neighbors})$, in which $|\m{E}_{sample}|$ is the number of edges in the sampled graph and $N_{neighbors}$ is the number of neighbors to be pruned.
Somewhat differently, for the Compressed Sparse Row (CSR) format, after deleting the edges, the row index arrays need to be adjusted accordingly, with prune complexity of $O(|\m{V}_{sample}|+N_{neighbors})$, where $|\m{V}_{sample}|$ is the number of nodes in the sampled graph.}



To reduce these graph pruning costs on GPU, we designed a novel data structure called \mcsr for \system. \mcsr uses two arrays to represent row indices -- the first array records the starting offset of a node's neighbors to the column index array while the second array records its ending offset. As illustrated in \Cref{fig:sparse-format}, for node $i$, its neighbors are stored in the column index segment starting from $start[i]$ to $end[i]$. To remove a node's neighbors, we can simply set the corresponding $end[i]=start[i]$ without any changes to the column index array, resulting in an $O(1)$ prune complexity. The data structure is also suitable for parallel processing on GPU because there is no data race condition.  

\section{Data Loader}
\label{sec:data-loader}

\margin{6-3}
\revision{Once a subgraph has been pruned, \system needs to load the historical embeddings for cached nodes and the features for unpruned nodes. Unlike historical embeddings stored in local GPU memory, features are stored on CPU (single-GPU training) or remote GPUs (multi-GPU training). Therefore, the loading process is still critical to performance.
}


However, the indices for unpruned nodes and their features reside on different devices. The index is calculated in the GPU (computation device) while the needed features are stored on the CPU or remote GPUs (storage device). 
Under such conditions, the naive way to fetch the features is via \textbf{two-sided} communication. This involves first transferring node indices from the computation device to the storage device, compacting the corresponding features on the storage device, and then sending the packed features back to the computation device.  This process introduces extra communication for transferring indices and synchronization between the computation and storage devices.
\margin{6-3}
\revision{The \system data loader can initialize memory access from a GPU side computation device (\textit{one-sided communication}) and change the communication schedule to fully utilize the bandwidth (\textit{multi-round communication}). As a result, it can efficiently load the features of the unpruned nodes.}

\begin{figure}[ht]
\centerline{\includegraphics[width=0.48\textwidth]{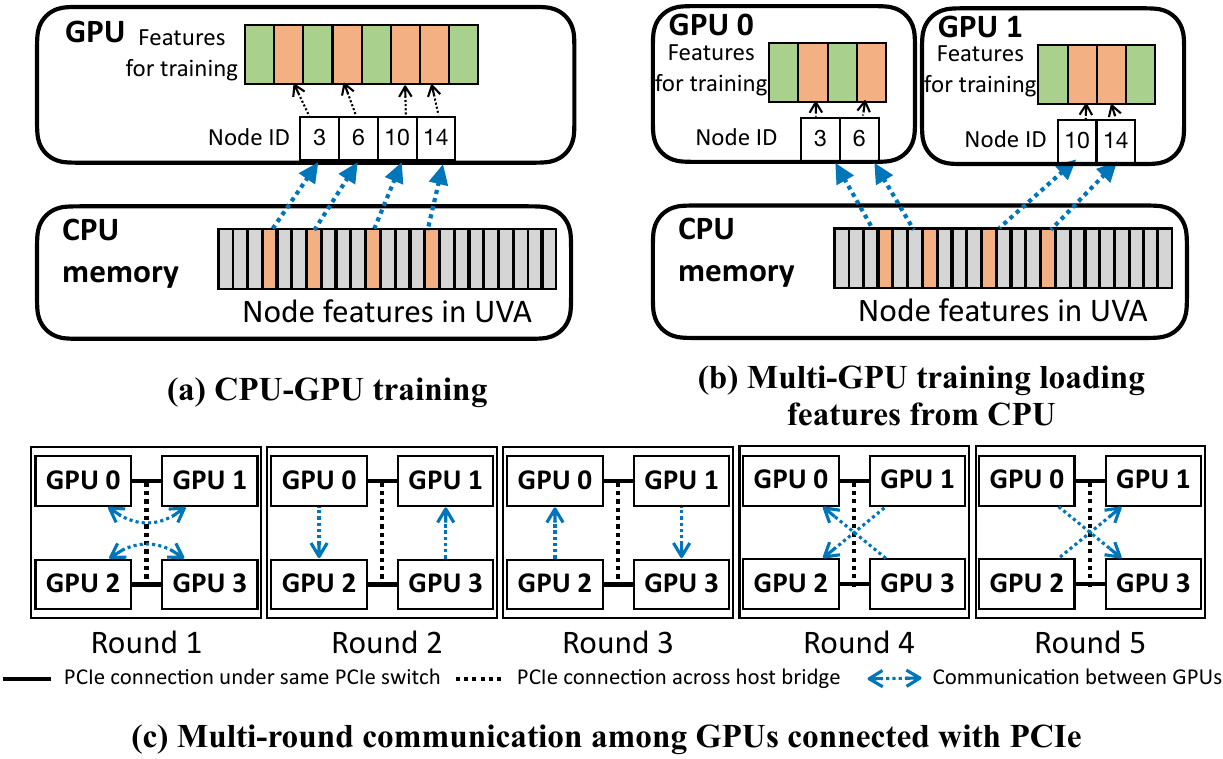}}
\rvspace{-5pt}
\caption{Feature data loading for (a) CPU-GPU training and (b) multi-GPU training.}
\label{fig:data-fetcher}
\end{figure}

\para{One-sided Communication.} \system employs one-sided communication to address this problem. 
Based on Unified Virtual Addressing (UVA)~\cite{pytorch-direct}, the CPU and GPU memories are mapped to a unified address.
This enables the computation device to directly fetch features from a mapped buffer of the storage device using the node index.
As shown in \Cref{fig:data-fetcher}(a), nodes needed for training are partly pruned by the cache (green color), and for the unpruned ones (orange color), the GPU fetches  features using their node ID directly from node features mapped with UVA. In \Cref{fig:data-fetcher}(b), multiple GPUs can concurrently fetch data using UVA for parallel training.


\para{Multi-round Communication.} With a larger number of GPUs available during training, all node features can be partitioned and stored across multiple machines such that GPUs serve as both computation and storage devices. Therefore, each GPU fetches the features of the relevant unpruned nodes from other GPUs, resulting in all-to-all communication between every pair of GPUs.
UVA can still be used to perform one-sided memory access in this scenario. However, as GPUs are connected asymmetrically, link congestion could badly degrade the overall bandwidth.
To address this problem, in addition to one-sided communication, \system breaks cross-GPU communication into multiple rounds to avoid congestion and fully utilize the bi-directional bandwidth on links. \Cref{fig:data-fetcher}(c) shows a typical interconnection among four GPUs, where GPUs are first connected via PCIe and then bridged via a host. For this topology, \system will decompose the all-to-all communication into five rounds. In round one, data is exchanged only between GPUs under the same PCIe switch, while the remaining four rounds are for exchanging data between GPUs across the host bridge. The data transmission at each round is bi-directional to fully utilize the bandwidth of the underlying hardware. This multi-round communication can effectively avoid congestion in all-to-all communication.

%% file: eval.tex
\section{Evaluation}
\label{sec:eval}

In this section, we first describe our experimental setup, followed by results covering system efficiency and model accuracy. We conclude with an empirical study of our cache effectiveness, ablations over our system optimizations, and heterogeneous extensions.

\subsection{Experimental Setup}

\begin{table}[ht]
\centering
\rvspace{-5pt}
\caption{Graph dataset details, including input node feature dimension (Dim.) and number of classes (\#Class).}
\rvspace{-10pt}
\label{tab:dataset}
\footnotesize
\begin{tabular}{lcccc}
\toprule
Dataset
& $|\m{V}|$  & $|\m{E}|$ & Dim.\tablefootnote{The first three datasets use float32 while the latter three use float16~\cite{ogb,ogb-lsc}.} & \#Class \\
\midrule

\texttt{ogbn-arxiv}~\cite{ogb} & 2.9M & 30.4M & 128 & 40 \\
\texttt{ogbn-products}~\cite{ogb} & 2.4M & 123M & 100 & 47 \\
\texttt{ogbn-papers100M}~\cite{ogb} & 111M & 1.6B & 128 & 172 \\
\texttt{MAG240M}~\cite{ogb-lsc} & 244.2M & 1.7B & 768 & 153 \\
\texttt{Twitter}~\cite{twitter} & 41.7M & 1.5B & 768 & 64 \\
\texttt{Friendster}~\cite{friendster} & 65.6M & 1.8B & 768 & 64 \\
\bottomrule
\end{tabular}
\end{table}

\para{Environments.}
Experiments were conducted on servers each with two AMD EPYC 7742 CPUs (2$\times$64 cores in total) and four NVIDIA A100 (40GB) GPUs connected via PCIe 4.0. The software environment on this machine is configured with Python v3.9, PyTorch v1.10, CUDA v11.3, DGL v0.9.1, and PyG v2.2.0.

\para{Datasets.} 
The dataset statistics are listed in \Cref{tab:dataset}. Among them, the two smallest datasets, \texttt{ogbn-arxiv} and \texttt{ogbn-products}, are included only for model accuracy comparisons and to provide contrast with much larger datasets, including \texttt{ogbn-papers100M} and \texttt{MAG240M}~\cite{ogb-lsc} that are used to test both accuracy and speed.  
Following the common practice of previous work~\cite{p3,gnnlab}, we also use the graph structure from 
\texttt{Twitter}~\cite{twitter} 
and 
\texttt{FriendSter}~\cite{friendster} 
with artificial features to test the speed of different systems.

\para{GNN models \& Training details.}
We employed three widely-used GNN architectures for our experiments: GraphSAGE~\cite{graphsage}, GCN~\cite{GCN}, and GAT~\cite{gat}. All models have 3 layers and 256 as the hidden size. 
\margin{4-5}
\revision{The base sampling method for mini-batch training is neighbor sampling, and we follow the setting of OGB leaderboard~\cite{ogb} to set the neighbor sampling fan-out as 20, 15, and 10.}
To measure their baseline model accuracy, we train the models using mini-batch neighbor sampling in DGL. 
The batch size is chosen to be 1000. We set $p_{grad}=0.9$ and $t_{stale}=200$ for \system for all experiments (with the exception of \Cref{sec:sensitivity}, where we study the impact of these thresholds).
\margin{4-4}
\revision{For speed tests, as the performance bottleneck is data loading, we only measure the performance of different systems on GraphSAGE; similar performance occurs with other models.}





\subsection{System Efficiency}
\label{sec:eval-perf}
We first demonstrate the system advantage of \system against state-of-the-art{/representative} alternatives for mini-batch training, including PyG~\cite{pyg}, GAS~\cite{gas}, ClusterGCN~\cite{clustergcn}, DGL~\cite{dgl} (as well as DistDGL~\cite{distdgl}), PyTorch-Direct~\cite{pytorch-direct}, MariusGNN~\cite{mariusgnn}, and GNNLab~\cite{gnnlab}.

\begin{figure}[ht]
    \centering
    \includegraphics[width=0.48\textwidth]{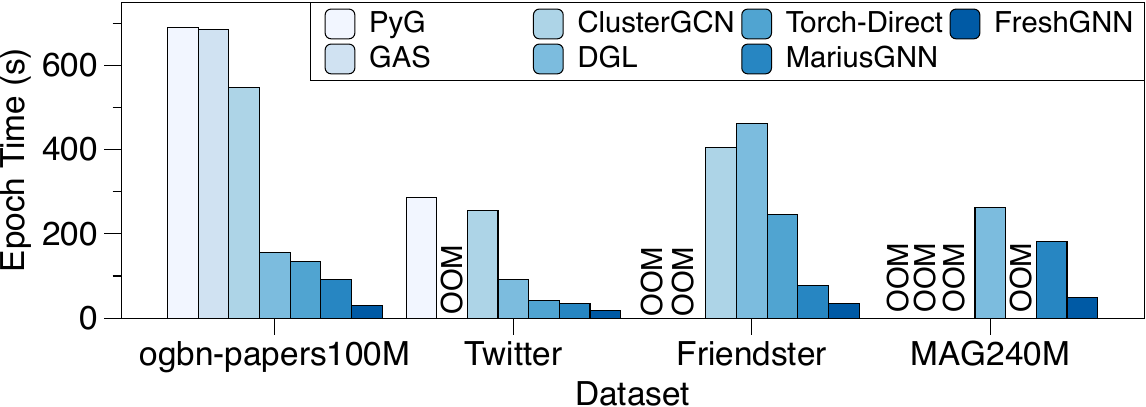}
\vspace{-20pt}
    \caption{Epoch time comparisons training a GraphSAGE model using a single GPU.}
\vspace{-10pt}
    \label{fig:overall-single-gpu}
\end{figure}


\Cref{fig:overall-single-gpu} compares the time for training a GraphSAGE model for one epoch on the four large-scale graph datasets using a single GPU. \system significantly outperforms all the other baselines across all the datasets. 
On \texttt{ogbn-papers100M} and \texttt{MAG240M}, \system has average speedup of 3.4$\times$ over the best of other baselines (MariusGNN).
Compared with the widely-used GNN systems DGL and PyG, \system is 5.3$\times$ and 23.6$\times$ faster respectively on \texttt{ogbn-papers100M}. Both PyTorch-Direct and \system utilize CUDA UVA memory to accelerate feature loading, but \system is still 4.6$\times$ faster because it can reduce the number of features to load by a large margin. With respect to other mini-batch training algorithms, \system is orders of magnitude faster than GAS and ClusterGCN on \texttt{ogbn-papers100M}. GAS also runs out of memory on graphs that have either more nodes/edges or larger feature dimensions due to the need to store the historical embeddings of all the nodes.  Incidentally, on the largest dataset \texttt{MAG240M} only DGL, MariusGNN, and \system avoid OOM, but with \system executing 5.3$\times$ faster than DGL and 3.7$\times$ faster than MariusGNN.


\begin{figure}[ht]
    \centering
    \rvspace{-5pt}
    \includegraphics[width=0.4\textwidth]{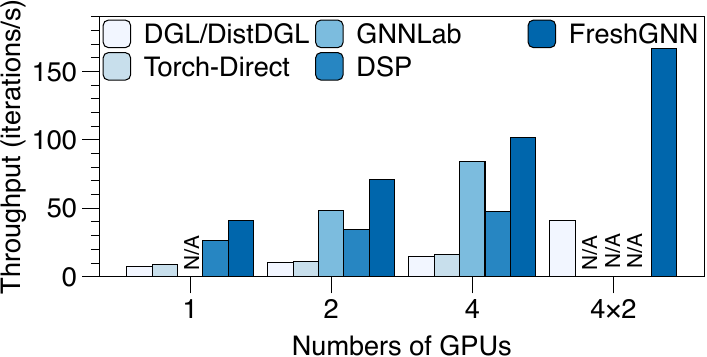}
\rvspace{-5pt}
    \caption{Scalability comparison for training a GraphSAGE model on \texttt{obgn-papers100M} using multiple GPUs.}
\rvspace{-5pt}
    \label{fig:overall-multi-gpu}
\end{figure}


\system can also scale to multiple GPUs and machines. In this setting, we include new baselines DistDGL~\cite{distdgl} for distributed training,
\margin{M-2, 5-1}
\revision{DSP~\cite{dsp} for fast multi-GPU sampling and training}, and GNNLab~\cite{gnnlab} which partitions GPUs to sampling or training workers; hence GNNLab does not have single-GPU performance.  
\Cref{fig:overall-multi-gpu} compares the throughput (measured as the number of iterations computed per second) when training GraphSAGE on \texttt{ogbn}-\texttt{papers100M}. Both DGL and PyTorch-Direct deliver almost no speedup because of the data loading bottleneck that cannot be parallelized via addition of more GPUs.
\margin{M-2, 5-1}
\revision{DSP utilizes GPU resources for graph sampling. However, as its graph structure is stored in a distributed manner on multiple GPUs, extra communication is needed among GPUs, which leads to inferior performance on our testing server equipped with PCIe connections.}
Meanwhile, \system enjoys good scalability from 1 to 4 GPUs and is up to {1.49$\times$} faster than GNNLab. When scaling to two machines each with four GPUs, the scalibility of \system becomes better, as it has more CPU resource for graph sampling, and is 4.07$\times$ faster than DistDGL.


\begin{figure}[ht]
\rvspace{-5pt}
\centerline{\includegraphics[width=0.4\textwidth]{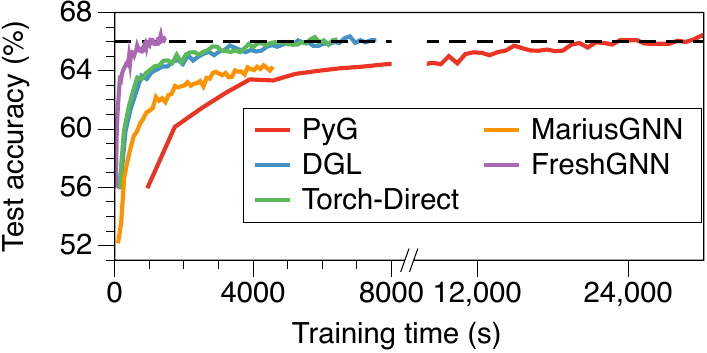}}
\rvspace{-5pt}
\caption{Test accuracy versus training time comparisons for GraphSAGE on \texttt{ogbn-papers100M}.}
\vspace{-5pt}
\label{fig:time-to-convergence}
\end{figure}

To further showcase \system speed advantages, \Cref{fig:time-to-convergence} plots the time-to-accuracy curve of different training systems {across 50 epochs}. 
{Except MariusGNN}, all the baselines here are using mini-batch neighbor sampling without any further approximation so they  converge to the same accuracy of $\sim$66\%; {MariusGNN obtains a lower accuracy of $\sim$63\% and further epochs did not improve performance}.
\system can reach this same accuracy in 25 minutes; the slowest baseline (PyG) takes more than 6 hours.


\begin{figure*}
    \centering
    \begin{subfigure}[b]{0.23\textwidth}
        \centering
		\includegraphics
		[width=\textwidth]{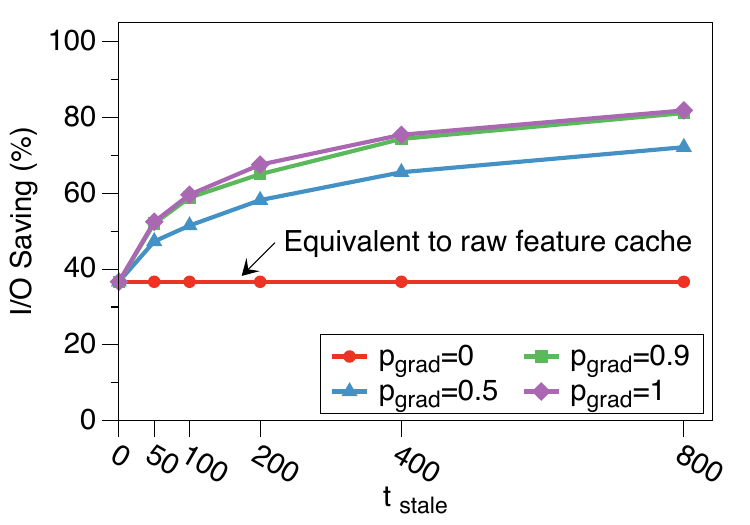}
		\vspace{-15pt}
	    \caption{\texttt{ogbn-papers100M}}
     \end{subfigure}
     \begin{subfigure}[b]{0.23\textwidth}
        \centering
		\includegraphics
		[width=\textwidth]{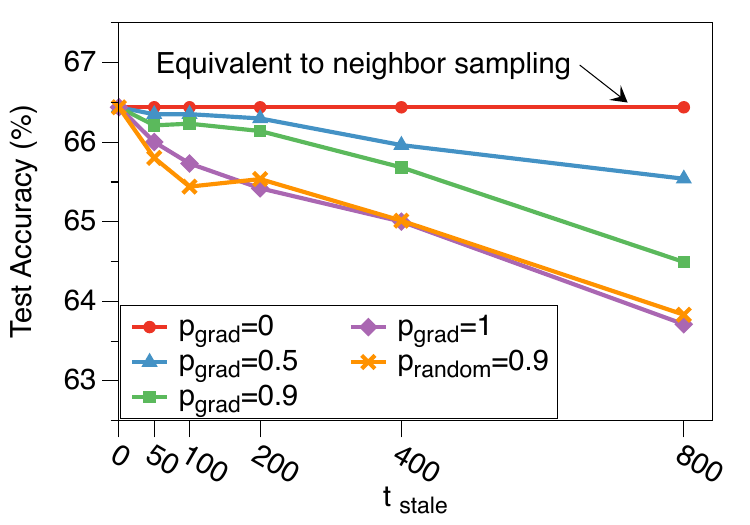}
		\vspace{-15pt}
		\caption{\texttt{ogbn-papers100M}}
     \end{subfigure}
     \begin{subfigure}[b]{0.23\textwidth}
        \centering
		\includegraphics
		[width=\textwidth]{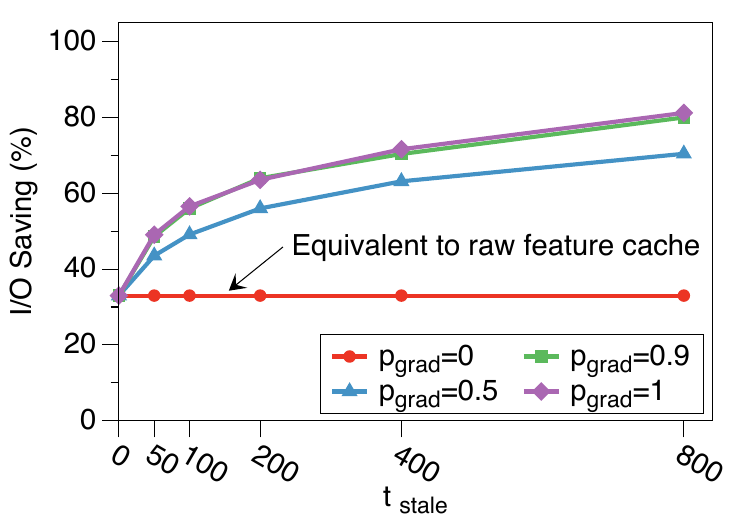}
		\vspace{-15pt}
		\caption{\texttt{MAG240M}}
     \end{subfigure}
     \begin{subfigure}[b]{0.23\textwidth}
        \centering
		\includegraphics
		[width=\textwidth]{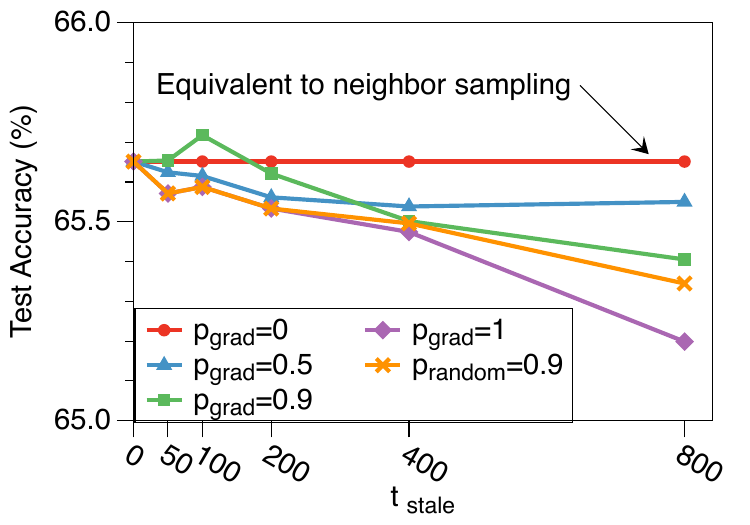}
		\vspace{-15pt}
		\caption{\texttt{MAG240M}}
    \end{subfigure}
\caption{(a) (c) The percentage saving of I/O for loading node features and (b) (d) the test accuracy on \texttt{ogbn-papers100M} and \texttt{MAG240M} under different choices of $p_{grad}$ and $t_{stale}$.}
\rvspace{-5pt}
\label{fig:cache-effectiveness}
\end{figure*}

\subsection{Model Accuracy}
\label{sec:eval-acc}

\Cref{tab:overall} compares the test accuracy of \system with other mini-batch training algorithms GAS~\cite{gas}, ClusterGCN~\cite{clustergcn}, GraphFM~\cite{graphfm}, and MariusGNN~\cite{mariusgnn}, as well as the full-graph training algorithm SANCUS~\cite{sancus} (which also uses historical-embeddings as described previously). Following the common practice in~\cite{gns, p3}, the target accuracy is obtained from training the base models (GraphSAGE, GAT, GCN) using neighbor sampling.



In general, nearly all algorithms perform relatively well on small graphs such as \texttt{ogbn-arxiv} and \texttt{ogbn-products} with only a few exceptions.
However, when scaling to larger graphs such as \texttt{ogbn-papers100M}, most of the baselines experience a substantial accuracy drop (from 7\% to 18\%) while running out of memory on \texttt{MAG240M}. {MariusGNN can run on all datasets with {GraphSAGE and GCN} (but not GAT), albeit with lower accuracy (over 2\% drop) and longer epoch time (from \Cref{fig:overall-single-gpu}).}
{We remark that SANCUS, as the representative of scalable full graph training, has low accuracy or suffers from OOM on large graphs due to the required storage overhead mentioned in \Cref{sec:history-full-graph}.}
By contrast, \system only experiences a less than 1\% accuracy difference across all datasets and base models.

\input{acctable.tex}

\subsection{Cache Effectiveness}
\label{sec:sensitivity}

Recall that $p_{grad}$ and $t_{stale}$ are the two thresholds that control the admission and eviction criteria of the historical embedding cache. In this section, we study their impact on system performance and model accuracy. We will show that a straightforward choice of the two thresholds can lead to the aforementioned competitive system speed as well as model accuracy.



\para{Impact on System Performance.}
In typical cache systems, the I/O saving percentage is equal to the cache hit rate. However, hitting the cached historical embeddings of a node will prune away all the I/O operations that would otherwise be needed to load the features of multi-hop neighbors, meaning the I/O savings can potentially be much larger than the cache hit rate. We plot this saving percentage w.r.t.~neighbor sampling (i.e., no caching) under different $p_{grad}$ and $t_{stale}$ in \Cref{fig:cache-effectiveness} (a) and (c). Because the historical embedding cache in \system is initialized via raw node features (\Cref{sec:gpu-implement}), the red $p_{grad}=0$ line reflects the performance of neighbor sampling with a raw feature cache.
As expected, larger $p_{grad}$ or $t_{stale}$ results in more I/O reduction. On both graph datasets, a raw feature cache can only reduce I/O by $<40\%$ but the historical embedding cache can reduce I/O by more than 60\% when choosing $t_{stale}>200$.

\begin{figure}[ht]
 \rvspace{-5pt}
\centerline{\includegraphics[width=0.4\textwidth]{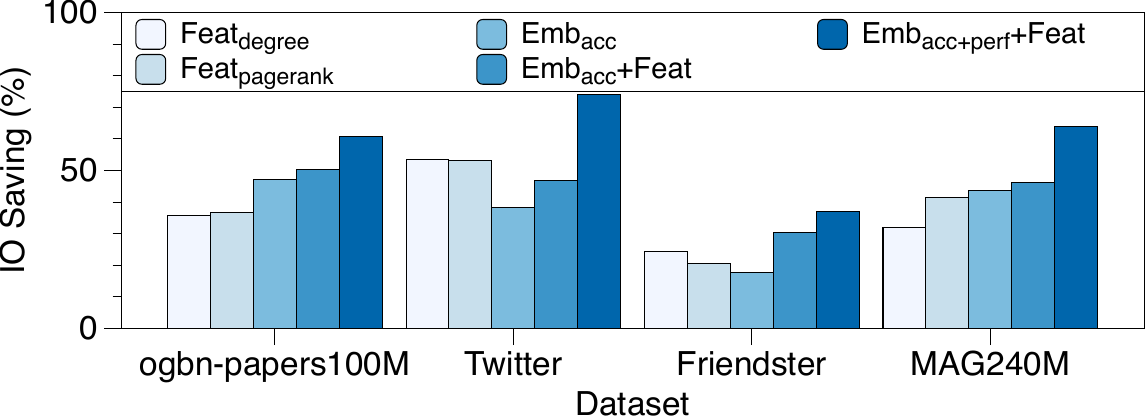}}
\rvspace{-7pt}
\caption{IO saving using different caching policies}
\rvspace{-5pt}
\label{fig:eval-io-saving}
\end{figure}

{We further present a detailed comparison of IO saving among various cache policies in \Cref{fig:eval-io-saving}. Specifically, we evaluate the degree-based and PageRank-based cache policies~\cite{gnntier} for feature cache and investigate three caching policies for historical embedding cache: cache policy for accuracy ($Emb_{acc}$), cache policy for accuracy with features ($Emb_{acc} + Feat$), and the cache policy for accuracy and performance with features ($Emb_{acc+perf} +Feat$) that utilizes bidirectional cache and prioritizes embeddings with high memory access savings. 
Our findings suggest that the IO saving achieved through feature caching is similar across different policies (on average 36.4\% vs. 38.0\%), as there are only a few hot nodes. On the other hand, caching historical embeddings alone does not result in significant IO saving (36.7\%), as the embedding cache policy is limited to stable embeddings. However, by combining feature and embedding caching, we observe a substantial improvement (43.4\%) compared to using either of the two caching schemes individually. Furthermore, by leveraging cache policy for performance to select embeddings that save a larger amount of memory access, \system achieves even higher IO savings (59.0\%).}


\para{Impact on Model Accuracy.}
We plot the corresponding accuracy curves under different configurations in \Cref{fig:cache-effectiveness} (b) and (d). As expected, larger $p_{grad}$ or $t_{stale}$ means more relaxed control on the embedding errors which consequently results in lower test accuracy. Beyond this, there are two interesting findings with practical significance. First, we can set $p_{grad}$ very close to one without an appreciable impact on the model accuracy. This aligns well with the observation in \Cref{fig:cos-simlarity-emb} that most of the node embeddings are temporally stable and can be safely admitted to the cache. Secondly, with a proper $p_{grad}$ value, GNN models can tolerate node embeddings that were last updated hundreds of iterations ago. By cross-checking the I/O saving figures, we find that this is also a sweet spot in terms of system performance, which eventually led us to choose $p_{grad}=0.9$ and $t_{stale}=200$ for all the experiments (obviously excluding the \Cref{fig:cache-effectiveness} results).

\revision{The accuracy of \system can be further improved using the  stabilization period from Section \ref{sec:instability}. }
\margin{4-2}
\revision{\Cref{tab:eval-cold-start} shows the accuracy of GraphSAGE with the historical cache introduced at different iterations (for reference, there are 1200 iterations in an epoch for \texttt{papers100M}). Here 0 means starting the historical cache at the beginning of training, and $INF$ means never starting historical cache, which is equivalent to regular mini-batch training. For \texttt{papers100M}, the effect is inconsequential, as the accuracy gap is small even without a stabilization period. However, for \texttt{mag240M}, the accuracy gap can be reduced from -0.51 to -0.12 with the stabilization period.}

\begin{table}[]
\caption{\revision{Accuracy with historical cache introduced at different iterations.  INF refers to never starting historical cache, which reproduces the original target accuracy.}}
\label{tab:eval-cold-start}
\footnotesize
\revision{
\begin{tabular}{@{}lccccc@{}}
\toprule
Iteration ID & 0     & 200   & 400   & 800   & INF/Target accuracy   \\ \midrule
\texttt{papers100M}   & -0.15 & -0.29 & -0.06 & -0.27 & 66.43 \\
\texttt{MAG240M}      & -0.51 & -0.12 & -0.18 & -0.12 & 66.14 \\ \bottomrule
\end{tabular}
}
\rvspace{-5pt}
\end{table}


\subsection{Ablation Study of System Optimizations}
\label{subsec:ablation}

In this section, we study the system performance of individual components. All the results are tested on \texttt{ogbn-papers100M} with a three-layer GraphSAGE model. 

\begin{figure}[ht]
\centerline{\includegraphics[width=0.4\textwidth]{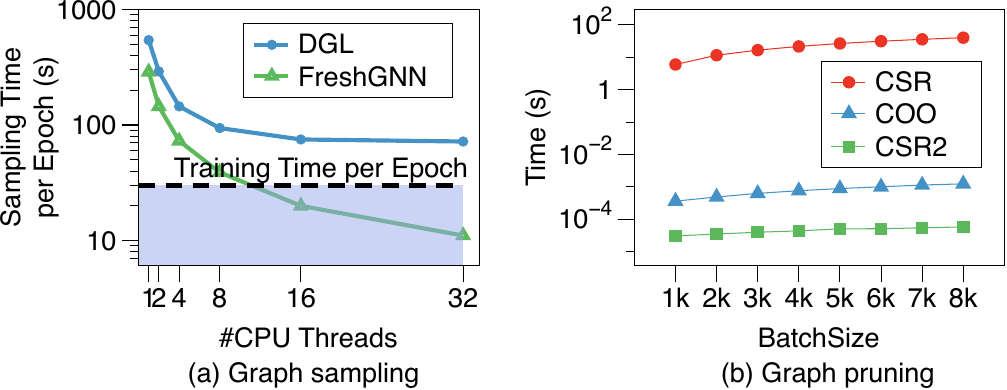}}
\caption{Effectiveness of \system's subgraph generator (measured on \texttt{ogbn-papers100M}).}
\label{fig:eval-generator}
\end{figure}

\para{Subgraph Generator.}
\revision{\Cref{fig:eval-generator}(a) shows sampling time per epoch of \system and DGL, as well as the training epoch time achieved by \system.}
\revision{\system sampler shows good scalability with more CPU threads.}
\margin{4-1, 4-6}
\revision{With 32 CPU threads, \system is able to reduce the sampling time per epoch to 11 seconds, which can be overlapped by training epoch time ($\sim$30s).} While DGL takes more than 70 seconds to finish graph sampling.

\Cref{fig:eval-generator}(b) measures the time to prune the cached nodes and their neighbors from a subgraph. Overall, \mcsr is orders-of-magnitude faster than CSR and COO regardless of the batch size in use. Specifically, CSR requires frequent CPU-GPU synchronization when invalidating the neighbors of the pruned nodes in the column index. As a result, graph pruning in total takes 99\% of the iteration time.  Subgraph pruning using COO is faster, which reduces the pruning overhead to 4.5\% of the iteration time, but is still much slower than \mcsr. The subgraph pruning time using \mcsr is negligible -- it only occupies 26$\upmu$s per iteration.

\begin{figure}[ht]
\centerline{\includegraphics[width=0.45\textwidth]{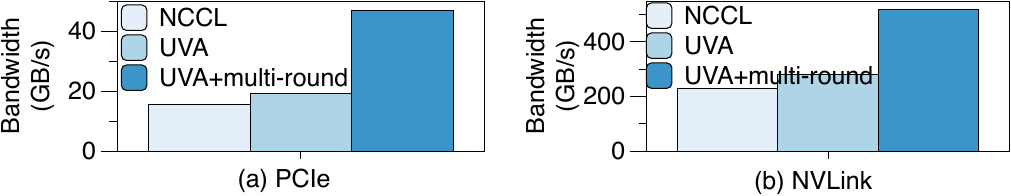}}
\rvspace{-5pt}
\caption{Optimizations for multi-GPU communication}
\rvspace{-5pt}
\label{fig:data-load-impv}
\end{figure}

\para{Data Loader.}
\Cref{fig:data-load-impv} shows the improvement afforded by the optimizations of \system's data loader for multi-GPU communication, on the aforementioned PCIe server and a server equipped with NVLink. Compared with the communication utilizing NCCL all-to-all operations, the one-sided communication is 23\% faster on average on PCIe and NVLink GPUs. After scheduling using the multi-round communication pattern, the bandwidth is increased by 145\% and 85\%, respectively.

\begin{figure}[ht]
\centerline{\includegraphics[width=0.45\textwidth]{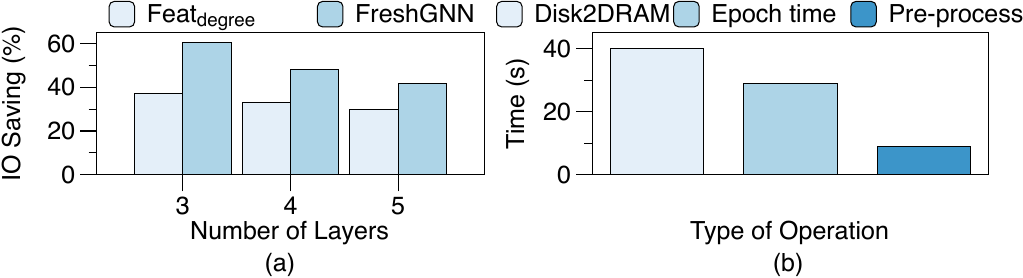}}
\caption{\revision{(a) IO savings for \texttt{papers100M} with varying model depth; (b) Overhead analysis for training on \texttt{papers100M} }}
\label{fig:eval-revision}
\end{figure}

\margin{6-2}
\revision{\para{IO Savings with Varying Model Depth}. \Cref{fig:eval-revision}(a) shows how IO savings change with GNN depth. For both feature cache and historical caches adopted by \system, the IO saving decreases with the increased number of layers and accessed nodes. However, \system consistently maintains higher IO saving of 1.50$\times$ compared to a standard feature cache based on node degree.}

\revision{
\para{Pre-processing Overhead.} In \Cref{fig:eval-revision}(b), we compared the pre-processing overhead with other necessary steps during training, such as loading node features from the disk (Disk2DRAM) and the training time for each epoch (Epoch time). The pre-processing overhead is smaller compared to these steps. Additionally, for the same dataset, the pre-processing results can be reused multiple times. Therefore, the pre-processing overhead of \system is not a significant factor.
}
\margin{4-3}

\begin{figure}[ht]
\centerline{\includegraphics[width=0.45\textwidth]{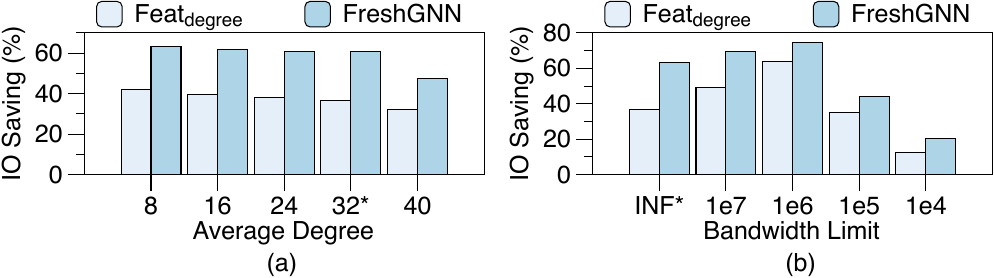}}
\caption{\revision{IO saving for (a). graphs with different average degrees; (b). graphs with different bandwidth limits. }}
\label{fig:eval-diverse-graph}
\end{figure}

\margin{5-2}
\revision{\para{IO Savings with Different Graph Topologies}. \Cref{fig:eval-diverse-graph} shows the IO saving from historical cache  on graph structures generated with different topology. By changing the average node degree and diameters, we find that historical cache used by \system can always lead to more IO saving than degree-based feature cache (1.56$\times$ and 1.43$\times$).}

\subsection{Extension to Heterogeneous Graphs}
\begin{figure}[h]
\centerline{\includegraphics[width=0.5\textwidth]{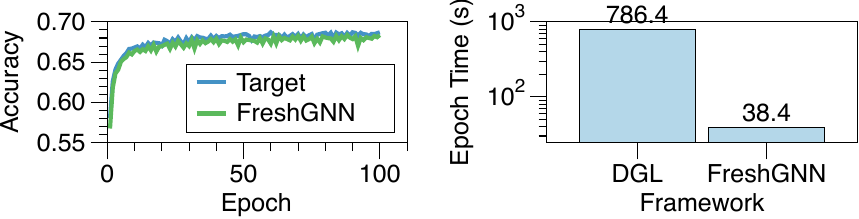}}
\caption{R-GCN evaluation on \texttt{MAG240M}.}
\label{fig:rgcn}
\end{figure}

While our primary focus has been on the most popular large-scale benchmark settings, \system can also be naturally extended to heterogeneous graphs. We now show one such use case on \texttt{MAG240M}, which was homogenized in the previous experiments. Here we instead use the heterogeneous form, along with R-GCN~\cite{rgcn}, which is arguably the most popular heterogeneous GNN architecture. 
In \Cref{fig:rgcn}, we compare \system with the neighbor sampling DGL baseline, {which is the only related work capable of running in this setting; other baselines either do not support heterogeneous graphs or else suffer from OOM.} From these results we observe that while the accuracy is almost the same, \system is 20.5$\times$ faster.

%% file: acctable.tex
\begin{table}[ht]
\caption{Test accuracy of different training algorithm minus the target accuracy obtained by neighbor sampling (larger is better). Bold numbers denote the best performing method.}
\rvspace{-10pt}
\footnotesize
\label{tab:overall}
\setlength{\tabcolsep}{4pt}
\begin{tabular}{ll|cc|cc}
\toprule[1.2pt]
&  & \multicolumn{2}{c|}{Small datasets} & \multicolumn{2}{c}{Large datasets} \\
\hline 
& Methods     & \texttt{arxiv} & \texttt{products} & \texttt{papers100M} & \texttt{MAG240M}
\tablefootnote{We report validation accuracy for MAG240M due to the absence of labels for test set.} \\
\midrule[1.2pt]
\multirow{7}{*}{\rotatebox{90}{\small GraphSAGE}}
& Target Accuracy & 70.91          & 78.66          & 66.43          & 66.14 \\ \cline{2-6}
&  GAS            & +0.44          & -1.19          & -8.17          & OOM  \\
& ClusterGCN      & -3.10          & +0.00          & -7.57          & OOM \\ 
& GraphFM         & \textbf{+0.62} & -7.90          & -18.40         & OOM \\
& MariusGNN         & -3.91  & -4.33         & -3.43         & -2.97   \\
& SANCUS         & -4.42  & -7.83         & OOM         & OOM   \\
& \system         & +0.60          & \textbf{+0.38} & \textbf{-0.15} & \textbf{-0.51}\\
\midrule[0.8pt]
\multirow{7}{*}{\rotatebox{90}{\small GAT}}
& Target Accuracy & 70.93          & 79.41          & 66.13          & 65.16 \\ \cline{2-6}
& GAS             & \textbf{-0.04} & -2.23          & -8.67          & OOM  \\
& ClusterGCN      & -3.91          & -2.92          & -8.08          & OOM  \\
& GraphFM         & -22.67         & -16.54         & OOM            & OOM   \\
& MariusGNN         & -1.37  & OOM          & OOM         & OOM   \\
& SANCUS         & -2.47  & OOM         & OOM         & OOM   \\
& \system         & -0.50          & \textbf{-0.54} & \textbf{-0.71} & \textbf{-0.36} \\
\midrule[0.8pt]
\multirow{7}{*}{\rotatebox{90}{\small GCN}}
& Target Accuracy & 71.24          & 78.57          & 65.78          & 65.24 \\ \cline{2-6}
& GAS             & +0.44          & -1.91          & -12.29         & OOM   \\
& ClusterGCN      & -3.13          & \textbf{+0.40} & -12.43         & OOM \\
& GraphFM         & \textbf{+0.47} & -15.70         & -18.70         & OOM   \\
& MariusGNN         & -0.55  & -1.10         & -2.04         & -2.37   \\
& SANCUS         & -3.04  & -9.44         & OOM         & OOM   \\
& \system         & -0.71          & -0.31          & \textbf{-0.16} & \textbf{-0.29} \\
\bottomrule[1.2pt]
\end{tabular}
\end{table}

%% file: related.tex
\section{Related Work}
\label{sec:related}

\para{Full-Graph GNN Systems.}
There exists a rich line of work on scaling full-graph GNN training, where the general idea is to split the graph into multiple partitions that can fit into the computing devices. Unlike mini-batch training, full-graph training updates all the nodes and edges simultaneously.
Notable systems include NeuGraph~\cite{neugraph}, ROC~\cite{ROC}, and DistGNN~\cite{distgnn}, which design smart data partitioning or data swap strategies to reduce the memory consumption and I/O cost of full-graph training. DGCL~\cite{dgcl} proposes a new communication pattern to reduce network congestion.
Dorylus~\cite{dorylus} and GNNAdvisor~\cite{gnnadvisor} optimize the computation of GNN training by exploring properties of the graph structure. BNS-GCN~\cite{bnsgcn} proposes boundary-node-sampling to reduce the communication incurred by boundary nodes during full graph training. 
{Beyond these, PipeGCN~\cite{pipegcn} and DIGEST~\cite{digest} utilize stale representations of neighbor nodes to reduce communication overhead, and MGG~\cite{mgg} applies pipelined communication for overhead hiding.} Finally, as discussed in \Cref{sec:history-full-graph}, {SANCUS~\cite{sancus} uses historical embeddings to reduce communication overhead among GPUs, and compares them with authentic embeddings to limit bias.}



\para{System Optimizations for GNNs.}
Previous work has shown that data movement is the bottleneck for GNN mini-batch training on large graphs. By using smart graph partitioning~\cite{metis,gnntier,distdgl,bgl} and data placement~\cite{p3,gnnlab,pytorch-direct,dsp}, this cost can be reduced under various training settings.
For example, DistDGL~\cite{distdgl} replicates high-degree nodes, together with sparse embedding updates, to reduce the communication workload for distributed training. 
\revision{DSP~\cite{dsp} stores node features and graph structure in a distributed way on GPUs to fully utilize NVlink bandwidth.}\margin{5-1} MariusGNN~\cite{mariusgnn} addresses the scenario of out-of-core training, reducing data swaps between disk and CPU memory by reordering training samples for better locality. \system's selective historical embedding method is orthogonal to both DistDGL and MariusGNN, and can potentially be combined with them. Other systems, such as GNNLab~\cite{gnnlab} and NextDoor~\cite{nextdoor}, utilize GPUs to accelerate the graph sampling process. These techniques are complementary to \system{} which optimizes feature loading.

\para{Training with Staleness.}
Staleness criteria have long been adopted for efficiency purposes when training deep neural networks.  Systems including SSP~\cite{ssp}, MXNet~\cite{parameter-server}, and Poseidon~\cite{poseidon} update model parameters asynchronously across different devices, helping to reduce global synchronization among devices and improve parallelism.
Another branch of work, including PipeDream~\cite{pipedream} and PipeSGD~\cite{pipesgd}, pipeline DNN training and utilize stale parameters to reduce pipeline stall. They limit the staleness by periodically performing global synchronization. 
\system, however, selectively stores and reuses stale node embeddings instead of parameters which is unique to GNN models.

\margin{6-1, M3}
\revision{\para{Cache for GNNs.}
There are two ways to cache for GNN training. The first is based on finding and caching node features that are frequently visited to save memory access. Previous methods utilize node degree~\cite{gnnadvisor,marius++,pagraph}, PageRank~\cite{gnntier,agnn}, or profiling~\cite{gnnlab} to estimate the frequency. 
The second is based on historical embeddings. Previous work~\cite{gas,vrgcn} caches either the historical embeddings of all nodes, or  boundary nodes \cite{sancus}. Such methods may experience out-of-memory or low accuracy on large graphs as we have shown.}

%% file: conclusion.tex
\section{Conclusion}

In this paper, we propose \system, a general framework for training GNNs on large, real-world graphs. At the core of our design is a new mini-batch training algorithm that leverages a historical cache for storing and reusing GNN node embeddings to avoid re-computation from raw features. To identify stable embeddings that can be cached, \system designates a cache policy using a combination of gradient-based and staleness criteria. Accompanied with other system optimizations, \system is able to accelerate the training speed of GNNs on large graphs by 3.4$\times$ over state-of-the-art systems, with less than 1\% influence on model accuracy.
